\documentclass[letterpaper,twocolumn,10pt]{article}
\usepackage{usenix}

\usepackage{amsmath}
\usepackage{amssymb}
\usepackage{tcolorbox}
\usepackage{xspace}
\usepackage{rotating}
\usepackage{tablefootnote}
\usepackage{pifont}
\usepackage{wasysym}
\usepackage{multirow}
\usepackage{makecell}
\usepackage{booktabs,caption}
\usepackage{changepage}
\usepackage[flushleft]{threeparttable}
\usepackage{color}
\usepackage{xcolor}
\usepackage{colortbl}
\usepackage{soul}
\usepackage{pifont}
\usepackage{comment}
\usepackage{pifont}
\usepackage{wrapfig,lipsum,booktabs}
\usepackage{subcaption}
\usepackage{tikz}
\usepackage{hyperref}
\usepackage{enumitem}

\usepackage{ulem}
\usepackage{xurl} % 比 url 包更智能的断行
\newcommand{\name}{\texttt{PatchHunter}\xspace}

\usepackage{filecontents}

\begin{document}
%-------------------------------------------------------------------------------

%don't want date printed
\date{}

\title{\Large \bf Pixel-Optimization-Free Patch Attack on Stereo Depth Estimation }

% for single author (just remove % characters)
\author{
% {\rm Anonymous author(s)}\\
  Hangcheng Liu\textsuperscript{1}, Xu Kuang\textsuperscript{2}, Xingshuo Han\textsuperscript{1}, Xingwan Wu\textsuperscript{3}, Haoran Ou\textsuperscript{1}, Shangwei Guo\textsuperscript{2},\\
   Xingyi Huang\textsuperscript{4}, Tao Xiang\textsuperscript{2}, Tianwei Zhang\textsuperscript{1}\\
  %   about author (webpage, alternative address)---\emph{not} for acknowledging
  %   funding agencies.} \\
  \small\textsuperscript{1}{Nanyang Technological University}, \textsuperscript{2}{Chongqing University}, \textsuperscript{3}{Wuhan University}, \textsuperscript{4}{Jinan University} \\
   % \textit{\{hangcheng.liu, tianwei.zhang\}@ntu.edu.sg}, \textit{haoran007@e.ntu.edu.sg}, \textit{\{xukuang, swguo, txiang\}@cqu.edu.cn}, \\
   % \textit{\{hanxingshuo, xyhuang81\}@gmail.com}, \textit{wxingwan@outlook.com}
} % end author

\maketitle

%-------------------------------------------------------------------------------
\begin{abstract}
%-------------------------------------------------------------------------------
Stereo Depth Estimation (SDE) is a core technique for scene perception in vision-based systems, particularly in autonomous driving. Recent studies have revealed that SDE is vulnerable to adversarial attacks based on pixel optimization. However, these attacks are confined to impractical settings, such as separately perturbing stereo views, restricted to the digital level, and static scenes, rendering them ineffective in real-world scenarios. A fundamental question remains: how can we design deployable, scene-adaptive, and transferable attacks against SDE models under realistic constraints?

We make two key contributions toward answering this question. First, we build a unified attack framework that extends existing pixel-optimization-based attack techniques to four key stages of stereo matching: feature extraction, cost-volume construction, cost aggregation, and disparity regression. Using this framework, we conduct a comprehensive, stage-wise evaluation of adversarial patches across 9 mainstream SDE models under realistic constraints (e.g., photometric consistency), showing that these attacks consistently suffer from poor transferability. Second, we present \name, a simple yet effective adversarial attack on SDE that operates without pixel-level optimization. \name formulates patch generation as a search task over a structured space of visual patterns designed to disrupt key SDE assumptions, and uses a reinforcement learning policy to discover effective and transferable patterns efficiently.

To comprehensively assess its effectiveness, transferability, and practicality, we evaluate \name across three levels, i.e., autonomous driving dataset, high-fidelity simulator, and real-world deployment. On the KITTI dataset, \name achieves superior attack effectiveness over pixel-level attacks, while significantly improving transferability under black-box scenarios. Further evaluations in the CARLA simulator and on a vehicle equipped with industrial-grade stereo cameras highlight its resilience to physical-world variations. Even under challenging conditions, such as low lighting, \name can achieve a D1-all error (ranging from 0 to 1) exceeding 0.4, whereas pixel-level attacks yield results close to 0.
\end{abstract}

\section{Introduction} \label{sec:intro}
Vision-based depth estimation infers scene depth information using RGB cameras, offering a low-cost and lightweight method in automated platforms such as autonomous driving~\cite{Menze2015CVPR,yang2019drivingstereo,PseudoLiDAR,autopilot,mobileye}. Existing vision-based methods can be categorized as Monocular Depth Estimation (MDE)~\cite{yang2024depth} and Stereo Depth Estimation (SDE)~\cite{cheng2025monster}. MDE estimates depth from a single camera image by learning semantic and geometric priors from data. While attractive for its minimal hardware requirements, MDE suffers from fundamental limitations, including scale ambiguity and poor generalization to unseen environments~\cite{ranftl2021vision,xian2020structure,bian2019unsupervised}. In contrast, SDE emulates human binocular vision by analyzing disparities between two images captured from horizontally displaced viewpoints. By leveraging explicit geometric constraints, SDE mitigates scale ambiguity and improves the accuracy of metric depth estimation~\cite{chang2018pyramid,smolyanskiy2018importance,tosi2019learning}.

Recent studies~\cite{wong2021stereopagnosia,berger2022stereoscopic,wang2024left,liu2024physical} have shown that SDE is susceptible to adversarial attacks, threatening its reliability in safety-critical applications. Existing works adopt pixel-optimization-based strategies, such as Projected Gradient Descent (PGD)~\cite{pgd}, which iteratively update pixel values via gradients to generate adversarial perturbations. By introducing these perturbations to the input images, these methods can significantly distort the predicted disparity maps, potentially leading to severe perception failures. Despite promising results in controlled settings, existing attacks rely on impractical assumptions, limiting real-world applicability.
Specifically, most works~\cite{wong2021stereopagnosia,berger2022stereoscopic,wang2024left} conduct digital-level attacks by applying different perturbations to the left and right views separately. This design violates the photometric consistency constraint inherent to stereo vision in the real world. Moreover, they typically require perturbations to be injected into the image streams in real time, making the physical-world deployment infeasible. Liu et al.~\cite{liu2024physical} enforce the photometric consistency and deploy their patch in the real world. However, since their attack is designed for static images, it cannot be applied to dynamic, real traffic scenarios.
% However, it is only evaluated on static images, leaving its effectiveness, transferability, and robustness unverified in dynamic, real traffic scenarios.

\begin{figure}[t]
    \centering
    \includegraphics[width=1.0\linewidth]{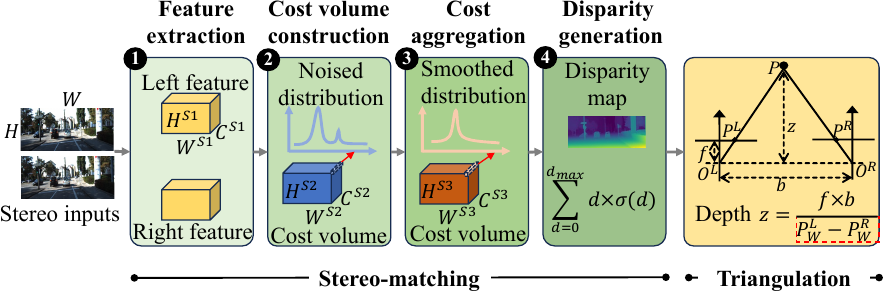}
    \caption{SDE comprises stereo matching and triangulation. The stereo matching (left) includes feature extraction (Stage~\ding{182}), cost volume construction (Stage~\ding{183}), cost aggregation (Stage~\ding{184}), and disparity estimation (Stage~\ding{185}), where $d_{max}$ is the maximum disparity and $\sigma(d)$ denotes the probability of disparity $d$. The resulting disparity map is converted to depth via triangulation (right) using focal length $f$ and baseline $b$.}
    \label{fig:fourstage}
\end{figure}

These limitations highlight the gap between controlled-setting attacks and practical, real-world deployment, raising two critical questions: \textit{Can pixel-optimization-based strategy still produce adversarial patches that are deployable, adaptive to various scenes, and transferable across diverse SDE models? If not, what other methods can effectively meet these requirements?} Motivated by these questions, our work delivers the following novel contributions:

\textbf{(1) We construct the first pixel-optimization-based adversarial attack framework that systematically generates adversarial patches by targeting the four key stages of stereo matching, enabling a comprehensive assessment of existing attack strategies and uncovering their poor transferability across diverse SDE architectures.}
This stage-wise attack design is motivated by the insight that the stereo matching module in SDE involves four interdependent stages: feature extraction, cost volume construction, cost aggregation, and disparity generation, which jointly determine the final depth output (Figure~\ref{fig:fourstage}). Disrupting any of these stages may lead to the failure of depth estimation. Our framework not only covers existing attacks, such as those targeting disparity generation for effectiveness~\cite{cheng2022revisiting,liu2024physical} and feature extraction for transferability~\cite{wang2024left}, but also identifies two previously underexplored stages, namely cost volume construction and cost aggregation, as new attack surfaces. Perturbing these intermediate stages can achieve comparable adversarial effects to those on the previously explored stages.

We comprehensively evaluate different attacks originating from this framework on the KITTI dataset~\cite{Menze2015CVPR} against nine representative SDE models. 
% This leads to two important findings. 
\textit{\ul{Evaluation results reveal that pixel-optimization-based attacks struggle to achieve high transferability under practical constraints such as photometric consistency}~\cite{cheng2022revisiting}\ul{ and scene adaptability}~\cite{berger2022stereoscopic}.}
Among 72 source-target model pairs formed from 9 SDE models, adversarial patches generated at the feature extraction, cost volume construction, cost aggregation, and disparity regression stages show poor transferability in 85\% (61 pairs), 92\% (66 pairs), 100\% (72 pairs) and 100\% (72 pairs) of the cases, respectively (Table~\ref{tab:percent}), indicating severe cross-model transferability limitations. Although attacks targeting feature extraction show relatively better transferability, the advantage is marginal, challenging the optimistic conclusions drawn in~\cite{wang2024left}. 

\textbf{(2) We propose \name, the first pixel-optimization-free adversarial attack framework, which efficiently searches for effective and transferable adversarial patterns within a structured pattern space specifically designed to disrupt shared SDE assumptions (e.g., neighboring pixels have smoothly varying disparities).} In our framework, all patterns directly determine the adversarial patch's visual appearance, without requiring additional pixel-level optimization or fine-tuning. To this end, we construct a Pattern Primitive Dictionary (PPD) that encodes 9 basic visual attributes with the potential to compromise SDE performance, including patch shape, hollowness, block repetition, brightness, color, texture type, frequency, orientation, and blur. Due to the combinatorial nature of these primitives and their multiple discrete choices, this PPD defines a large and semantically meaningful pattern space with approximately 44,000 unique configurations. To sufficiently identify effective and transferable patterns from such a large space, we adopt a reinforcement learning (RL)–based search strategy that operates without accessing the victim SDE model's gradients.

Using \name, we identify a set of adversarial patterns for all tested SDE models on KITTI under two practical black-box settings (Section~\ref{sec:threatmodel}), both of which require no knowledge about the model's internal details. Our evaluations of these discovered patterns yield two outcomes.
\ul{\textit{1) Adversarial patches generated by}} \name \ul{\textit{exhibit clear superiority over pixel-optimization-based patches in terms of attack effectiveness and transferability.}} In a query-based black-box setting (\textit{interactive}), our patches achieve effectiveness that matches or surpasses the best results reported by pixel-optimization-based methods under white-box conditions, where they need to access the internal details of target SDE models, e.g., parameters. In a transfer-only black-box setting (\textit{non-interactive}), \name successfully breaks source-to-target pairs that previous methods consistently fail to attack, effectively raising the upper bound on transferability. Moreover, all evaluations are performed on KITTI images that were not used during the pattern search phase, confirming the discovered patterns generalize well to new, unseen scenes. \ul{\textit{2) The shared characteristics across all discovered patterns converge to an important adversarial configuration defined by high-density hollowness, block repetition, and high texture clarity.}} 
This convergence suggests the existence of an architecture-agnostic adversarial structure, which underpins the strong transferability observed in the non-interactive setting. Ablation studies show that altering any of these three attributes leads to a noticeable drop in attack performance, reinforcing the importance of this configuration. We attribute its effectiveness to the distinct, complementary, and mutually reinforcing disruptive effects introduced by each component.

\textbf{(3) We comprehensively evaluate \name on a high-fidelity autonomous driving simulator (CARLA) and a real Unmanned Ground Vehicle (UGV), demonstrating its strong practical applicability.} CARLA provides photorealistic scenes with fine-grained control over environmental factors, allowing us to systematically evaluate the attack effectiveness, transferability, and robustness across diverse scenarios, such as variations in distance, lighting, weather, patch rotation, and even camera configurations (e.g., stereo baselines). Additionally, we test \name using a UGV equipped with industrial binocular cameras for more practical evaluations. Results show that  \name \ul{\textit{consistently outperforms pixel-optimization-based attacks in the real world, even in challenging scenarios such as low-lighting conditions.}} More importantly, our patches are inherently robust without relying on additional techniques such as Expectation over Transformation (EoT)~\cite{athalye2018synthesizing}, highlighting the simple yet effective design of \name.

% More importantly, the robustness of our pattern-driven patches is inherent, without relying on techniques such as Expectation over Transformation (EoT)~\cite{athalye2018synthesizing}, highlighting the simple yet effective design of \name.

% Overall, \name demonstrates that a simple yet effective design centered on structured patterns can outperform pixel-optimization methods in both simulation and the real world.

\section{Stereo Depth Estimation} \label{sec:background:sde}

SDE relies on two principles: \textit{stereo matching} and \textit{triangulation}. Stereo-matching identifies 2D projections of the same 3D point in the left and right views, producing a disparity map, where each disparity value denotes the horizontal pixel offset between two projections. Triangulation then computes the 3D depth from this disparity using stereo camera parameters.

\subsection{Stereo Matching} 
Mainstream methods generally adopt a four-stage scheme, as illustrated in Figure~\ref{fig:fourstage}. 
\ding{182}~\textit{Feature extraction}. The stereo images are processed independently through a feature extractor to produce corresponding feature maps. Early methods employ residual networks~\cite{shaked2017improved} to construct the feature extractor, while later approaches introduce expanded receptive fields~\cite{park2016look} and multi-scale pyramid modules~\cite{chang2018pyramid} to enhance contextual understanding. \ding{183}~\textit{Cost volume construction}. This stage computes a cost volume, a multi-dimensional array that stores the matching cost for every pixel at every possible disparity. A typical cost volume has the shape $B\times C^{S2} \times H^{S2} \times W^{S2}$, corresponding to the dimensions of batch size, disparity level, height, and width, respectively. Applying a softmax along the disparity dimension can produce the disparity probability distribution for each location. Since the initial cost volume is often noisy and unreliable, it requires further refinement in the next stage. \ding{184}~\textit{Cost aggregation}. This stage integrates local and global information across the initial cost volume to suppress noise in disparity distributions. Most methods employ 3D convolutional networks~\cite{kendall2017end,chang2018pyramid} to regularize cost volumes across spatial and disparity dimensions. More recent methods explore efficient alternatives, such as guided aggregation~\cite{zhang2019ga} and adaptive aggregation~\cite{xu2020aanet}. \ding{185}~\textit{Disparity generation}. The refined cost volume is converted to a probability distribution with softmax over disparity, then averaged to produce the final disparity estimate. Yet, predictions can still be inaccurate in occlusion regions, fine structures, or textureless areas. Many methods incorporate refinement modules that leverage high-resolution features or edge-aware guidance to correct local errors~\cite{lipson2021raft, xu2023unifying}. These four interdependent stages not only determine the overall accuracy of SDE, but also constitute distinct attack surfaces. In Section~\ref{sec:empirical}, we will leverage this stage-wise structure to evaluate pixel-optimization-based attacks systematically.

\subsection{Triangulation}
Triangulation is the fundamental geometric principle that enables the recovery of 3D structure from stereo image pairs, converting pixel-level disparity measurements into metric depth estimates based on known camera intrinsics and extrinsics. As shown on the right of Figure~\ref{fig:fourstage}, triangulation can be formally expressed as
\begin{equation}
% \begin{split}
    z =\frac{f\times b}{d}, \text{where} \; d = P^L_W-P^R_W. \label{eq:stereo}
% \end{split}
\end{equation}
where $P^L_W$ and $P^R_W$ are the horizontal coordinates of the projections $P_L$ and $P_R$, and $d$ is their disparity. Both $f$ (focal length) and $b$ (baseline) are known camera parameters. Therefore, the accuracy of depth calculation directly relies on the quality of $d$. Consequently, existing adversarial attacks~\cite{wong2021stereopagnosia,berger2022stereoscopic,liu2024physical,wang2024left} on SDE focus on disrupting the stereo matching process. By default, attacks on SDE in the remainder of this paper refer to those targeting the stereo matching module.

% Unless otherwise specified, attacks on SDE discussed in the remainder of this paper are assumed to target the stereo matching.
% In the remainder of this paper, we refer to attacks on SDE as attacks targeting the stereo matching module.

\section{Pixel-optimization-based Attack Framework} \label{sec:empirical}

As mentioned in Section~\ref{sec:intro}, existing pixel-optimization-based SDE attacks~\cite{wong2021stereopagnosia,berger2022stereoscopic,wang2024left,liu2024physical} typically rely on impractical assumptions.
None of them can be simultaneously (i) physically deployable under unconstrained outdoor conditions, (ii) resilient to dynamic scenes, and (iii) transferable across diverse SDE backbones. This naturally raises a critical question: \textit{``Are pixel-level attacks truly practical, even with advanced enhancements?''}

To thoroughly answer this, we construct the first \ul{U}niversal \ul{S}tereo-\ul{C}onstrained Pixel-level Attack Framework (USC) against SDE. USC unifies prior strengths to satisfy real-world deployment needs and explores previously overlooked attack surfaces. Specifically, prior attacks mainly target feature extraction and disparity generation (stage \ding{182} and \ding{185}), neglecting the two equally critical intermediate stages of cost volume construction and cost aggregation (stage \ding{183} and \ding{184}). Cost volume construction encodes the initial disparity distribution by computing matching costs across potential disparities, while cost aggregation serves as a crucial bridge between raw matching costs and high-quality disparity estimates by context modeling and spatial regularization. Though largely overlooked, these two stages expose distinct vulnerabilities and represent promising attack surfaces. By targeting all four stages of stereo matching, USC enables a comprehensive evaluation of pixel-optimization-based attack strategies for SDE.

\subsection{Attack Framework Design} \label{sec:uscpgd}

In USC, we formulate adversarial patch generation for each stage as the following optimization problem:
\begin{equation}
        \min_{\mathbf{P}}\, \mathbb{E}_{(I^L,I^R)\sim \mathcal{D}}[\mathcal{L}^i_{\theta_{s}}(\hat{I}^L, \hat{I}^R)],
    \label{eq:usc}
\end{equation}
where 
\begin{equation}
\begin{aligned}
    \hat{I}^L[h, w] &=
    \begin{cases}
        \mathbf{\hat{P}}[h, w], & \text{if } (h, w) \in \Omega, \\
        I^L[h, w], & \text{otherwise},
    \end{cases} \\
    \hat{I}^R[h, w] &=
    \begin{cases}
        \mathbf{\hat{P}}[h, w + \mathbf{d}[h, w]], & \text{if } (h, w + \mathbf{d}[h, w]) \in \Omega, \\
        I^R[h, w], & \text{otherwise}.
    \end{cases}
\end{aligned}
\label{eq:constrains}
\end{equation}
Eq.~(\ref{eq:usc}) uses a stage-specific objective $\mathcal{L}^i_{\theta_s}$ to attack the $i$-th stage, where $\hat{I}^L$ and $\hat{I}^R$ represent the attacked left and right views by a patch $\mathbf{P}$. Each objective corrupts the intermediate output of the targeted stage to disrupt subsequent processing. 

Additionally, some practicality enhancements are involved in Eq.~(\ref{eq:usc}) and Eq.~(\ref{eq:constrains}). First, Eq.~(\ref{eq:usc}) minimizes the expected loss over natural stereo distributions $\mathcal{D}$ rather than a fixed benign input pair $(I^L, I^R)$, improving effectiveness under dynamic scenes~\cite{berger2022stereoscopic}. Second, Eq.~(\ref{eq:constrains}) enforces stereo-consistency constraints, meaning that the patch appears at geometrically corresponding positions in the left and right views as determined by the real disparity, thereby preserving real-world deployability~\cite{liu2024physical}. In this formulation, $\mathbf{\hat{P}}$ is a perturbation map with the same size as the input image, where only a designated region contains the patch $\mathbf{P}$ and all other pixels are zero. Below we elaborate $\mathcal{L}^i_{\theta_s}$ for attacking each stage.

\noindent\textbf{I. Attacking feature extraction.} Inspired by~\cite{wang2024left}, attacks targeting the feature extraction stage aim to reduce the similarity between the patch’s feature representations in the left and right views. The objective function is defined as:
\begin{equation}
    \mathcal{L}^1_{\theta_s}=\frac{1}{\|\Omega\|}\sum_{[h,w]\in \Omega } Cosine(F^L[h,w-\mathbf{d_s}[h,w]], F^R[h,w]).
\end{equation}
$F^E$ and $F^R$ are the left and right input features. $Cosine$ is the cosine similarity. $\mathbf{d_s}$ is the downscaled disparity $\mathbf{d}$, ensuring coordinate alignment with the lower-resolution feature maps.

\begin{figure*}
    \centering
    \includegraphics[width=\linewidth]{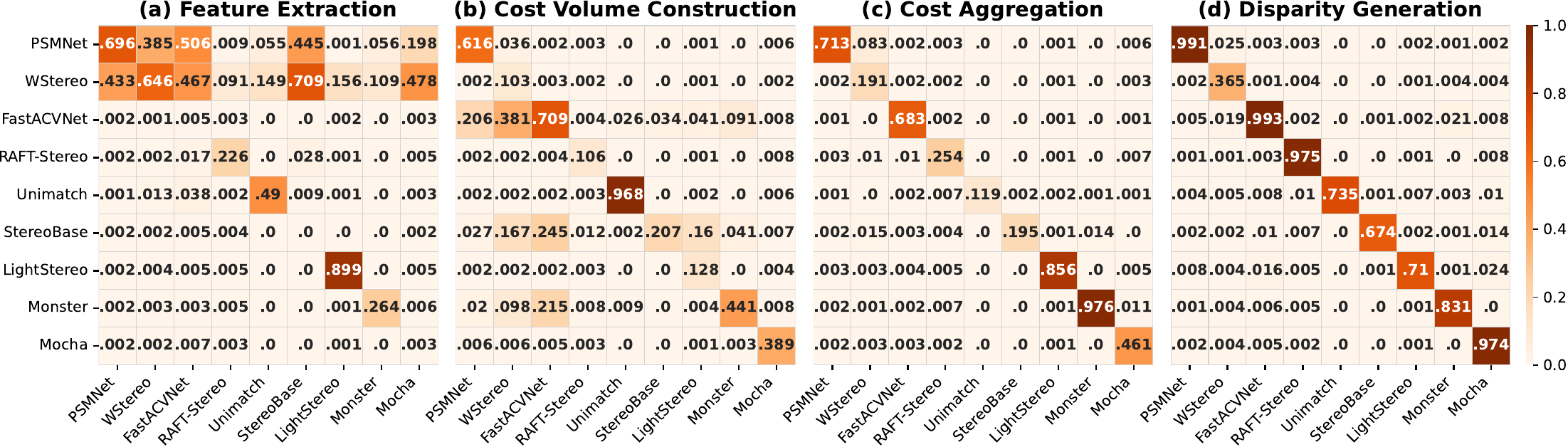}
    \caption{Stage-wise attack performance of USC on the four stereo matching stages, measured by D1-all (decimal part only). Each heatmap shows the attack results for one stage, with rows as surrogate models and columns as target models. Diagonals indicate white-box attacks; off-diagonals show black-box transfer. Darker cells indicate stronger attacks.}
    \label{fig:usc}
\end{figure*}

\noindent\textbf{II. Attacking cost volume construction.} Since no prior work has explored this stage, we design a new attack that injects noise into the cost volume. This is motivated by the fact that the disparity distribution in the initial cost volume $V$ directly influences subsequent refinement steps such as disparity smoothing. We define the objective function as:
\begin{equation}
    \mathcal{L}^2_{\theta_s}=\frac{1}{\|\Omega\|}\sum_{[h,w]\in \Omega } JSD(softmax(\mathbf{V}[h,w]),\mathcal{M}),
    \label{eq:loss2}
\end{equation}
where $JSD$ is the Jensen-Shannon Divergence, $softmax(\mathbf{V}[h,w])$ represents the normalized disparity distribution at location $[h,w]$, and $\mathcal{M}$ is a synthetic distribution with a explicit multi-modal structure (e.g., sine wave in Figure~\ref{fig:synthetic}). The design of this multi-modal distribution is motivated by two factors. First, ideal disparity distributions concentrate most of their probability mass around the true disparity. Enforcing a multi-modal structure disperses this energy, reducing the dominance of the correct match and increasing ambiguity. Second, Garg et al.~\cite{garg2020wasserstein} showed that multi-modal disparity may lead to significant disparity drift.

\noindent\textbf{III. Attacking cost aggregation.} Since both the initial and aggregated cost volumes represent per-pixel disparity distributions, manipulating their distributional structure can be achieved using the same divergence-based objective. Therefore, we reuse the objective function from the cost volume construction stage to attack cost aggregation, encouraging the aggregated volume $\mathbf{V'}$ to retain a multi-modal structure:
\begin{equation}
    \mathcal{L}^3_\theta=\frac{1}{\|\Omega\|}\sum_{[h,w]\in \Omega } JSD(softmax(\mathbf{V'}[h,w]),\mathcal{M}).
    \label{eq:loss3}
\end{equation}

\noindent\textbf{IV. Attacking disparity generation.} As the most commonly targeted stage in prior work, disparity generation is attacked by maximizing the error between predicted and ground-truth disparities. Following~\cite{liu2024physical}, we design the objective as:
\begin{equation}
    \mathcal{L}^4_{\theta_s}= -\frac{1}{\|\Omega\|}\sum_{[h,w]\in\Omega} \left | \mathbf{\tilde{d}}[h,w] - \mathbf{d}[h,w]  \right |^2.  
\end{equation}

Notably, USC not only explores new attack surfaces but also serves as a practicality-enhanced variant of existing attack methods~\cite{wong2021stereopagnosia,berger2022stereoscopic,wang2024left,liu2024physical}. It addresses both dynamic scene adaptability and deployability, whereas prior works consider only one of these aspects.

\subsection{Evaluation Setup} \label{sec:investigation:setup}

\noindent\textbf{Dataset.}  We evaluate our USC on the KITTI~\cite{Menze2015CVPR} benchmark. The KITTI-Stereo-2015 dataset contains 200 stereo pairs for training and 200 for testing. We use the full training set for patch optimization and the full test set for evaluation. The images span across diverse traffic scenes, including urban roads and rural paths. All images are zero-padded to the size of $384 \times 1248$ to meet model input requirements.

\noindent\textbf{SDE Models.} We select 9 SOTA SDE models with diverse architectures for evaluation: PSMNet~\cite{chang2018pyramid}, WStereo~\cite{garg2020wasserstein}, FastACVNet~\cite{xu2022attention}, RAFT-Stereo~\cite{lipson2021raft}, Unimatch~\cite{xu2023unifying}, StereoBase~\cite{OpenStereo}, LightStereo~\cite{guo2025lightstereo}, Monster~\cite{cheng2025monster}, and Mocha~\cite{chen2024mocha}. All models are evaluated using their official checkpoints trained on the KITTI dataset.

\noindent\textbf{Implementation.} We use Adam to optimize Eq.~(\ref{eq:usc}) with a learning rate of 0.004 for up to 3000 iterations. All pixel values are normalized to $[0, 1]$. The default patch size is $150\times 300$. We also consider other patch sizes in Appendix~\ref{sec:appendix:patchsize}.

\noindent\textbf{Evaluation Metrics.} 
We mainly use two metrics to evaluate attack performance. The first is D1-all, a widely used metric in SDE studies~\cite{wong2021stereopagnosia,berger2022stereoscopic}. It measures the percentage of pixels with disparity errors exceeding both 3 pixels and 5\% of the ground truth, which are considered to be erroneously estimated by the model. Formally, D1-all is computed as:
\begin{equation}
    \begin{split}
    \text{D1-all}&=\frac{1}{\left \| \Omega \right \| } \sum_{[h,w] \in \Omega } l[h,w],\\
        l[h,w]&=\left\{\begin{matrix}
 1 & \delta[h,w]>\max\{3,0.05\cdot \mathbf{d}[h,w]\},  \\
 0 & \text{otherwise},  
\end{matrix}\right.\\
 \delta[h,w] &=  \left | \mathbf{\tilde{d}}[h,w] - \mathbf{d}[h,w] \right |, 
    \end{split}
    \label{eq:d1all}
\end{equation}  
where $\mathbf{d}$ and $\mathbf{\tilde{d}}$ represent the ground-truth and predicted disparity maps. $[h,w]$ are specific height and width coordinates within the map, $\Omega$ denotes the set of all pixel coordinates, and $\|\Omega\|$ is its size. Since we focus on adversarial patch attacks, $\Omega$ is limited to pixels within the patch. D1-all (bounded between 0 and 1) reflects the \textit{pixel-level Attack Success Rate (ASR)}. Higher values indicate stronger attacks.

The second metric is $\mathcal{P}(\circ,\tau)$, introduced in this work to quantify the transferability performance between models.
\begin{equation}
    \mathcal{P}(\circ,\tau)=\frac{1}{|\mathbf{M}|(|\mathbf{M}|-1)}\sum_{\substack{F_s,F_t\in\mathbf{M}\\F_s \ne F_t}} \mathbb{I}(\text{D1-all}_{F_s\rightarrow F_t} \circ \tau), \label{eq:crasr}
\end{equation}
where $\circ$ is a relational operator ($<$ or $>$), $\tau$ is a threshold, $\mathbf{M}$ is the set of SDE models, $|\mathbf{M}|$ is the set size, $F_s$ is the surrogate model, $F_t$ is the target model, $\text{D1-all}_{F_s\rightarrow F_t}$ denotes the D1-all error when a patch generated for $F_s$ is applied to $F_t$. $\mathbb{I}$ is an indicator function outputting 1 if the condition holds and 0 otherwise. $\mathcal{P}(\circ, \tau)$ ranges from 0 to 1. Smaller $\mathcal{P}(<, \tau)$ indicates stronger worst-case transferability under threshold $\tau$, whereas a larger $\mathcal{P}(>, \tau)$ reflects stronger best-case transferability, equivalent to the \textit{cross-model-level ASR} under threshold $\tau$. Besides these two, we consider another popular metric End-Point Error (EPE), as detailed in Appendix~\ref{sec:appendix:epe}.

\begin{figure}[t]
    \centering
    \includegraphics[width=\linewidth]{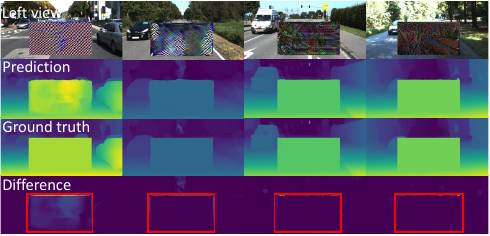}
    \caption{Poor transferability of USC. From left to right, the source-target pairs are (PSMNet, FastACVNet), (RAFT-Stereo, Unimatch), (Monster, RAFT-Stereo), and (Mocha, LightStereo). The difference maps show absolute disparity errors between the ground truth and predictions by the target models, with brighter areas indicating larger deviations. The deviations within the red boxes in the last three columns are imperceptible. All visualizations are
    cropped for clarity}
    \label{fig:fail_example}
\end{figure}

\subsection{Evaluation Results} \label{sec:answer}

We use USC to attack each stage of every target model and generate the corresponding adversarial patch. The effectiveness and transferability of each patch are then evaluated across all models. The results are shown in Figure~\ref{fig:usc}, from which we highlight the following key observations.

\textbf{(1) All four stage-wise attacks are effective under white-box dataset simulation.} As shown in Figure~\ref{fig:usc}, attacks on cost volume construction and aggregation, newly explored by USC, achieve comparable performance to the previously studied stages, validating the stage-wise design. Most diagonal cells show high D1-all errors, and later-stage attacks (e.g., cost aggregation, disparity regression) are generally more effective, likely because downstream modules refine and thus suppress early-stage perturbations.

\begin{figure}[t]
    \centering
    \includegraphics[width=\linewidth]{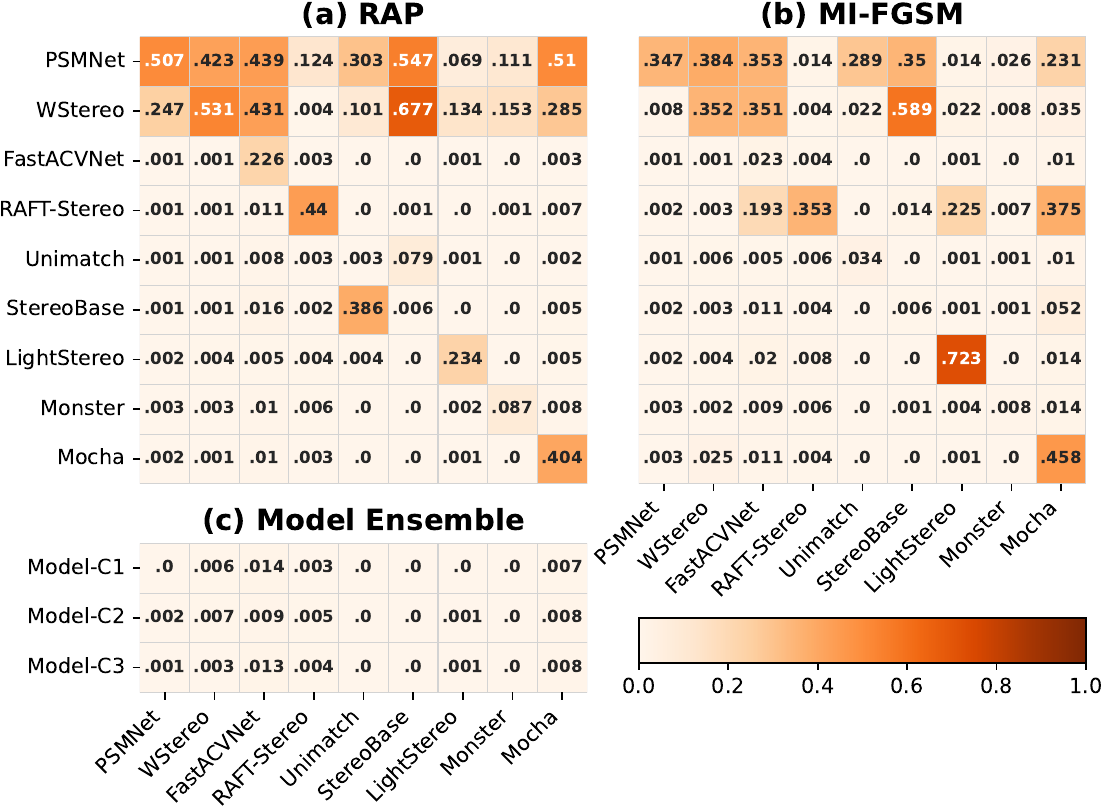}
    \caption{Existing transferability enhancement methods fail to improve USC. All D1-all values are obtained from attacks targeting the feature extraction stage. Specifically, Model-C1, Model-C2, and Model-C3 represent different surrogate model combinations. Model-C1: PSMNet, WStereo, and RAFT-Stereo. Model-C2: FastACVNet, Monster, and LightStereo. Model-C3: WStereo, Unimatch, and StereoBase.}
    \label{fig:existing-enhancement}
    % \vspace{-10pt}
\end{figure}

\textbf{(2) All four stage-wise attacks exhibit negligible or very limited transferability.} As illustrated in Figure~\ref{fig:usc}, $\mathcal{P}(<,0.1)$ stays above 0.85 across all four stages, suggesting poor transferability. Figure~\ref{fig:fail_example} shows some failure cases. We further apply three representative transferability enhancement methods to USC: RAP~\cite{qin2022boosting}, MI-FGSM~\cite{dong2018boosting}, and model ensemble~\cite{tramer2017ensemble}, all well-studied in other domains. RAP and MI-FGSM aim to improve the optimization trajectory, while the model ensemble increases surrogate diversity. We integrate them into USC when attacking feature extraction, which shows the highest transferability in Figure~\ref{fig:usc}. Figure~\ref{fig:existing-enhancement} shows that none of them yield meaningful transferability improvements. Notably, despite its success elsewhere, model ensemble offers no clear benefit, consistent with observations in~\cite{liu2024beware} for MDE attacks. We also consider attacking all four stages jointly, but this way still fails to improve transferability (Appendix~\ref{sec:appendix:multistage}). All results demonstrate that existing attack strategies struggle to achieve high transferability under realistic constraints.

\begin{figure}[t]
    \centering
    \includegraphics[width=\linewidth]{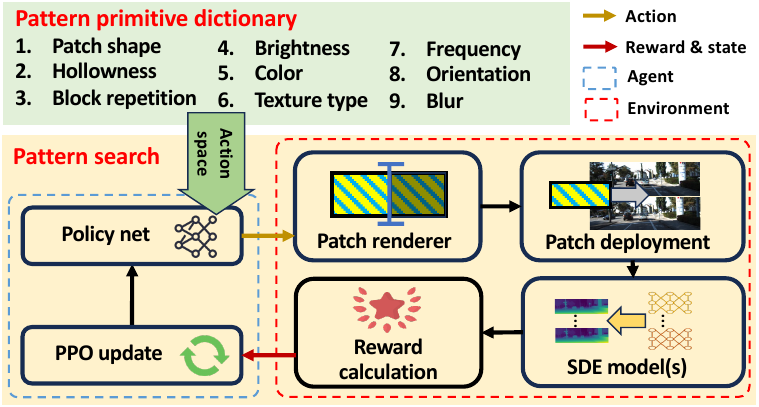}
    \caption{Overview of \name. }
    \label{fig:framework}
\end{figure}
\section{Methodology}
We present a simple yet effective attack methodology, \name,  to address the above-mentioned limitations of existing pixel-optimization-based attacks. 

\begin{table*}[t]
  \centering

  \caption{PPD comprises 4 categories covering 9 pattern primitives. Each primitive has multiple specific pattern options.}
    \resizebox*{\linewidth}{!}{
    \newcolumntype{L}[1]{>{\raggedright\arraybackslash}p{#1}}
    \begin{tabular}{p{3.665em}|p{3.415em}|p{2.085em}|p{2.415em}|p{11.185em}|p{12.585em}|p{3.385em}|p{4.085em}|p{.165em}}
    \hline
    \multicolumn{3}{c|}{Geometry} & \multicolumn{2}{c|}{Photometry} & \multicolumn{3}{c|}{Texture} & \multicolumn{1}{c}{Degradation} \\
    \hline
    \multicolumn{1}{c|}{Shape} & \multicolumn{1}{c|}{Hollowness} & \multicolumn{1}{c|}{\makecell{Block\\repetition}} & \multicolumn{1}{c|}{Brightness} & \multicolumn{1}{c|}{Color} & \multicolumn{1}{c|}{Texture type} & \multicolumn{1}{c|}{Frequency} & \multicolumn{1}{c|}{Orientation} & \multicolumn{1}{c}{Blur} \\
    \hline
    \makecell[tl]{Retangle;\\ Circle} & \makecell[tl]{No;\\Low density;\\High density} & No; Yes & Normal; Low; High & \makecell[tl]{B\&W; R\&G; Bl\&Y;\\W\&G; Y\&B; G\&P;\\LG\&MG; SB\&LC; O\&WC} & \makecell[tl]{Solid; Gradient; Chessboard;\\ Stripe; Wave; Concentric;\\ Zabra; Leopard; Random} & Low; High & \makecell[tl]{$\rightarrow$ $\leftarrow$ $\uparrow$ $\downarrow$ \\ $\nearrow$ $\searrow$ $\swarrow$ $\nwarrow$}      & \makecell[tl]{No; Slight; \\ Heavy} \\
    \hline
    \end{tabular}%
    }
\begin{adjustwidth}{0.5cm}{0.5cm}
\begin{tablenotes}[para,flushleft]
\footnotesize
	\centering \textbf{B}: Black\;\textbf{W}: White\; \textbf{Bl}: Blue\; \textbf{R}: Red\; \textbf{G}: Green\; \textbf{Y}: Yellow\; \textbf{P}: Purple\; \textbf{LG}: Light Green\; \textbf{MG}: Mint Green\; \textbf{SB}: Sky Blue\; \textbf{LC}: Light Cyan\; \textbf{O}: Orange\; \textbf{WC}: Warm Coral\;
\end{tablenotes}
\end{adjustwidth}
\label{tab:primitive}%
\end{table*}%

\begin{figure*}[t]
    \centering
    \includegraphics[width=\linewidth]{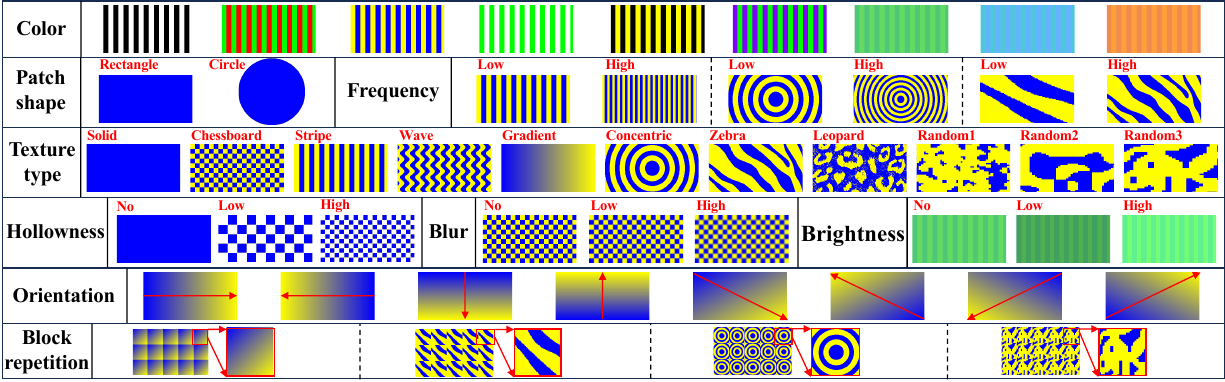}
    \caption{Visual examples of pattern primitives and their options used in our patch design.}
    \label{fig:opition}
\end{figure*}

\subsection{Threat model} \label{sec:threatmodel}
We adopt a widely used and realistic threat scenario explored in prior work~\cite{wei2024physical}, in which an attacker physically places a printable adversarial patch in the environment to disrupt the SDE function in autonomous driving scenarios.

\noindent\textbf{Attack goal.} Our goal is to make the SDE model's predictions within the patch region deviate significantly from the ground truth, similar to the objectives of prior works~\cite{wang2024left,berger2022stereoscopic}. The greater the deviation, the more effective the attack, as it increases the likelihood of inducing severe depth perception errors, potentially leading to safety-critical failures. 

\noindent\textbf{Adversarial capability.} We consider two practical black-box attack settings. 
\textit{(1) Interactive attack.} The adversary has no knowledge of the victim SDE model's internal information, but can query the model with his input and observe the corresponding output. This assumption is realistic in autonomous driving scenarios, where attackers may exploit interface-level vulnerabilities or debugging channels to intercept perception data. For example, in a documented incident in 2021, a hacker successfully extracted the input-output data stream from a Tesla vehicle’s depth estimation module~\cite{web2}. This suggests that an adversary could leverage an identical vehicle to probe the model's input-output behavior and then generate the adversarial patch based on the feedback. 
\textit{(2) Non-interactive attack.} In this more restrictive setting, the adversary has no access to the victim model’s inputs, outputs, or internal mechanisms. Thus, the patch generation can only be performed on a surrogate model, and the resulting patch must exhibit high transferability to succeed across unseen models.

\subsection{Key Insights and Challenges}

\textbf{Instead of optimizing patch pixels, \name aims to discover effective natural patterns that can break the \textit{fundamental assumptions} of SDE}. Regardless of architecture, SDE models inherently rely on assumptions such as spatial smoothness and photometric consistency. Certain visual patterns could challenge these architecture-agnostic assumptions, creating opportunities to induce errors across diverse models. However, identifying such patch patterns under black-box settings, especially those that consistently transfer across SDE models, is non-trivial. It requires a rich, structured pattern space that enables systematic exploration. To the best of our knowledge, no existing work has provided such a space or conducted a comprehensive study of diverse patterns, leaving a critical gap in understanding and exploiting transferable adversarial patterns. Motivated by the these, we construct the first structured pattern space composed of architecture-agnostic failure modes and systematically search it to identify adversarial patterns with high transferability.

%Motivated by this insight, 

% However, identifying effective patch patterns under black-box settings, especially those that consistently transfer across SDE models, is non-trivial. It requires a rich, structured pattern space that supports systematic exploration. To the best of our knowledge, no existing work has provided such a space or conducted a comprehensive study of diverse patterns, leaving a critical gap in understanding and exploiting transferable adversarial patterns.
% We construct the first structured pattern space composed of architecture-agnostic failure modes and systematically search it to identify adversarial patterns with high transferability.

\subsection{Overview} \label{sec:method:overview}
{Figure~\ref{fig:framework} shows the pipeline of \name, which transitions the attack paradigm from pixel-level optimization to pattern search. The central contribution of \name is the construction of the Pattern Primitive Dictionary (PPD), a structured space representation that enables systematic exploration of transferable adversarial patterns. The PPD organizes diverse low-level texture patterns into 9 basic, composable primitives (Table~\ref{tab:primitive}) that define common visual attributes such as patch shape, color, and texture type. Each primitive contains one or more specific options derived from commonly used configurations, which may challenge SDE models from multiple perspectives. The composability of primitives, coupled with a rich set of concrete pattern options, results in a large and expressive pattern space. To navigate this space and find transferable patterns, we adopt a Reinforcement Learning (RL)-based search strategy. RL has proven effective in prior work~\cite{yang2020patchattack} for exploring discrete adversarial configurations (patch textures and locations) in classification attacks, and we extend it to SDE attacks. As discussed in Appendix~\ref{sec:appendix:ga}, other search strategies, such as genetic algorithms, can also be applied, as our constructed PPD is algorithm-agnostic.

\subsection{Pattern Primitive Dictionary Construction} \label{sec:method:pattern}
As shown in Table~\ref{tab:primitive}, PPD defines our pattern space, structured into four semantic categories and nine basic pattern primitives. Figure~\ref{fig:opition} illustrates these pattern primitives and their specific options, providing an intuitive understanding of them. 
% Each primitive captures a distinct aspect of visual appearance and contributes to the compositional flexibility of the pattern space. 
The design of PPD and the specific options within each primitive are empirically motivated. In the following, we describe each primitive and its potential to challenge SDE models.
% , drawing from prior findings~\cite{schrodi2022towards,chang2018pyramid,Sharma_2018_ECCV}, commonly observed visual characteristics, and our targeted analysis of weaknesses in stereo matching pipelines. 

% \begin{packeditemize}
\textbf{(1) Geometric attribute.} Geometric primitives are designed to disrupt the spatial structural assumptions commonly leveraged by stereo matching, including geometric consistency, uniqueness, and completeness. \ding{172} \textbf{Patch shape} defines the global silhouette (e.g., rectangles or circles), affecting perceived object boundaries. \ding{173} \textbf{Hollowness} controls whether structured cutout regions appear within the patch and determines their scale. These cutouts fragment the spatial continuity and challenge the model’s ability to perform disparity smoothing. \ding{174} \textbf{Block repetition} determines whether identical texture blocks are repeatedly arranged across the patch, similar to a tiled floor. If enabled, it introduces visually repetitive patterns at the patch level, even when the texture within each block is irregular. This repetition introduces multiple look-alike regions, causing the stereo matching process to struggle in identifying which region in one view corresponds to the correct location in the other.  

\textbf{(2) Photometric attribute.} \ding{175} \textbf{Brightness} adjusts luminance to high, normal, or low levels, where extreme luminance may reduce local contrast and hinder feature extraction. \ding{176}~\textbf{Color} defines color composition using high- or low-contrast pairs. High contrast (e.g., black and white) creates strong edges and false structures, while low contrast (e.g., light cyan and mint green) suppresses saliency, leading to low-texture ambiguity.

\textbf{(3) Texture attribute.} \ding{177} \textbf{Texture type} determines the internal structure: (a) solid or gradient textures lack distinctive features; (b) regular textures (e.g., stripes, checkerboards) introduce global texture similarity, inducing ambiguity in stereo matching like block repetition; (c) irregular textures (e.g., random noise) typically exhibit high-frequency and incoherent structures, which may overly activate early-stage features in CNN-based models, leading to a noisy and unstable cost volume. \ding{178} \textbf{Frequency} controls the density of texture distribution within the patch. High-frequency settings may enhance the effects of regular or repetitive textures, while low-frequency settings can exacerbate the ambiguity associated with low-texture regions. \ding{179} \textbf{Orientation} specifies the dominant direction (e.g., horizontal, vertical, diagonal) of texture variation, i.e., the direction along which visual intensity changes occur as illustrated in Figure~\ref{fig:opition}. Table~\ref{tab:orientation} in the Appendix shows that changing the orientation could noticeably affect attack effectiveness even with all other patterns fixed.

\textbf{(4) Degradation attribute.} \ding{180} \textbf{Blur} controls the sharpness of texture boundaries by applying different types of blurring operations to the patch. A heavy blur reduces feature clarity, thereby having the potential to impair the localization and matching of local features. Conversely, a sharp and well-defined texture boundary can amplify the perceptual salience of other disruptive attributes, such as repeated textures, thereby jointly increasing the adversarial impact. 
    
% \end{packeditemize}
The implementations of these patterns can be found in Appendix~\ref{sec:appendix:patchhunter}. 
Notably, our PPD is both \textit{\textbf{diverse}} and \textit{\textbf{extendable.}}
It currently consists of nine basic pattern primitives, already defining an expressive and large space. After removing redundant configurations (e.g., the combination of different frequencies with solid color), PPD still yields nearly \textbf{44,000} unique and semantically meaningful pattern configurations, sufficient to produce effective adversarial patterns that outperform pixel-optimization-based patches in both transferability and real-world robustness (Section~\ref{sec:exp}). Moreover, users can incorporate additional custom patterns into the space by defining new primitives or extending existing ones, making the framework flexible for broader deployment scenarios.

\begin{table*}[t]
\centering
\footnotesize
\caption{Adversarial patterns discovered by \name for each SDE model and their D1-all scores (effecitiveness). ``---'' marks attributes not applicable to certain textures.}
\resizebox*{\linewidth}{!}{
\begin{tabular}{c|ccccccccc|c}
\hline
{Target model} & {Patch shape} & {Hollowness} & {Block repetition} & {Brightness} & {Color} & {Texture type} & {Frequency} & {Orientation} & {Blur} & {Effectiveness} \\
\hline
PSMNet       & Circle   & High density & Yes & Normal  & White\&Green             & Zebra     & Low    & --- &  No    & 0.842 \\
WStereo      & Rectangle& High density & Yes & High    & Red\&Green               & Gradient  & --- & $\searrow$   & Slight & 0.825 \\
FastACVNet   & Circle& High density & Yes & High & White                & Solid     & --- & --- & No     & 0.860 \\
RAFT-Stereo  & Rectangle& High density & Yes & Normal    & Black\&white           & Leopard     & High   & $\searrow$ & No     & 0.858 \\
Unimatch     & Circle& High density           & No & Normal     & Black & Solid      & ---    & ---      & No     & 0.620 \\
StereoBase   & Rectangle& No           & Yes & Normal  & White\&Green            & Wave      & Low    & $\rightarrow$        & No     & 0.799 \\
LightStereo  & Rectangle& High density & Yes & High    & Sky blue\&Light Cyan    & Gradient  & --- & $\uparrow$     & No     & 0.796 \\
Monster      & Rectangle& High density & Yes & Normal  & White                   & Solid     & --- & --- & No & 0.748 \\
Mocha        & Circle & No           & Yes & Normal  & Blue\&Yellow            & Wave      & High   & $\rightarrow$        & No     & 0.896 \\
\hline
\end{tabular}
}
\label{tab:pattern-decomposition}
\end{table*}

\subsection{Reinforcement Learning-based Search} 
The search process is modeled as an agent–environment interaction, where the agent receives feedback and selects one option per primitive to form a complete adversarial configuration. This way supports black-box search, making it well-suited for real-world attack settings.

\subsubsection{Environment}
The environment encapsulates the whole pattern configuration space and provides an interface for patch generation and evaluation. It is formalized by the RL tuple $(\mathcal{S}, \mathcal{A}, \mathcal{R})$, representing the state, action, and reward, respectively.

% \begin{packeditemize}
\noindent\textbf{State.}
We define the state $s_t\in\mathcal{S}$ at time step $t$ as the pattern configuration selected in the previous step, i.e., $s_t = a_{t-1}$. This design allows the agent to condition its current decision on its most recent output, enabling a form of short-term memory that encourages diverse exploration.

\noindent\textbf{Action.}
To enable policy-based exploration over the pattern space, we define the agent’s action $a_t$ at time $t$ as the selection of a complete pattern configuration. Specifically, given a set of $N$ pattern primitives ${\mathcal{O}_1, \mathcal{O}_2, ..., \mathcal{O}_N}$, where each $\mathcal{O}_i = {o_i^1, o_i^2, ..., o_i^{K_i}}$ is a finite option set with cardinality $K_i$, the action space is defined as:
\begin{equation}
        \mathcal{A} = \{(o_1^{k_1},o_2^{k_2}, \cdots, o_N^{k_N})\,|\,o_i^{k_i} \in \mathcal{O}_i\}.
\end{equation}
Each $a_t \in \mathcal{A}$ corresponds to a specific combination of visual primitives and is used to generate a candidate patch.

\noindent\textbf{Reward.}
Upon taking an action (pattern configuration), the environment generates an adversarial patch (patch render), applies it onto randomly sampled stereo images (patch deployment), and evaluates the performance degradation of SDE, quantified by D1-all. The D1-all score serves as the reward: $r_t=\text{D1-all}(a_t)$, indicating the attack effectiveness of the current configuration. Since the evaluation only requires access to the model's inputs and outputs, it supports the interactive setting. For the more restricted non-interactive setting, a surrogate model or model ensembles can be used to estimate the reward and enhance transferability.
% \end{packeditemize}

\subsubsection{Agent} The agent uses a policy network to select actions based on the current state: $\pi_{\theta_P}(a_t \mid s_t)$, where $\theta_P$ denotes all learnable parameters of the policy. No information from stereo images is used as input to the agent, avoiding overfitting to scene-specific content and encouraging the discovery of patterns with intrinsic transferability. We adopt a standard Proximal Policy Optimization (PPO) algorithm~\cite{schulman2017proximal} to train the policy network within 800 environment interaction steps (i.e., complete observation–action–reward cycles). The network architecture and training procedure details are provided in Appendix~\ref{sec:appendix:patchhunter}.

In subsequent evaluations, we use the optimal pattern combination found during agent training to define the final adversarial patch. Although this deviates from standard RL practice, where agents continue to interact with the environment after training, it aligns with our research objective. Rather than learning an ongoing decision policy, we aim to discover a robust universal pattern configuration across diverse scenes and transferable to various SDEs.

% We implement the policy network as a multi-head (e.g., 9 heads) Multi-Layer Perceptron (MLP), where each head corresponds to one pattern primitive and outputs a categorical distribution over its discrete options. We apply greedy decoding during inference by selecting the most probable option from each head to construct the complete action vector. This design enables the agent to generate a complete pattern configuration in a single forward pass. 

% \subsubsection{Training}
% We train the policy network using a standard Proximal Policy Optimization (PPO)~\cite{schulman2017proximal} algorithm with a clipped surrogate objective and normalized advantage, aiming to discover pattern configurations that maximize the reward (refer to Appendix~\ref{sec:appendix:patchhunter} for details).

% In our experiments, the policy network converges within just 800 environment interaction steps (Figure~\ref{fig:ppo_curves}), each corresponding to a complete observation–action–reward cycle. This is remarkably efficient given the large pattern space of nearly 44,000 unique configurations. Such efficiency stems from two key factors: (i) the highly structured pattern space defined by our PPD, which consists of semantically meaningful pattern options that have the potential to challenge various assumptions of SDE, and (ii) the simple yet effective design of our RL-based search procedure, which enables stable and fast convergence. 

\section{Experiments} \label{sec:exp}
We use \name to discover effective patterns for all evaluated SDE models on the KITTI dataset~\cite{Menze2015CVPR}. We evaluate all discovered patterns across three levels: a real-world autonomous driving dataset, a high-fidelity simulator, and physical-world deployment.

\subsection{Dataset Evaluation}
We first evaluate the \textit{effectiveness} and \textit{transferability} of \name's discovered patterns on KITTI to allow direct comparison with prior works.

\noindent\textbf{Setup.} Following the same experimental setup as described in Section~\ref{sec:empirical}, we use the training set of KITTI-Stereo-2015 to perform pattern search, and evaluate the discovered patches on the testing set. We maintain the patch-to-image area ratio (approximately 9.4\%) as used in Section~\ref{sec:investigation:setup}.

\noindent\textbf{Evaluations under \textit{interactive} attack.} In this setting, we directly use the victim SDE model's outputs to compute rewards during the pattern search, aiming to verify the attack effectiveness. Table~\ref{tab:pattern-decomposition} summarizes the patterns identified by \name for each victim SDE model, along with their corresponding effectiveness measured by D1-all on KITTI. These results reveal two findings: \ding{172} \ul{\textit{Our adversarial patches built from basic patterns can significantly degrade SDE performance in black-box settings}}, as shown by high D1-all scores ranging from 0.62 to 0.896.
\ding{173} \ul{\textit{All discovered patterns share common traits, high-density hollowness, block repetition, and non-blur}}, indicating these features are key to attacking SDE. Their recurrence across models suggests shared structural vulnerabilities, enabling high transferability despite architectural differences. To further validate the importance of high-density hollowness, block repetition, and non-blur, we conduct ablation experiments on PSMNet, RAFT-Stereo, and LightStereo. We take each model's original pattern configuration from Table~\ref{tab:pattern-decomposition} and vary one attribute at a time while keeping the other two fixed. As shown in Figure~\ref{fig:individual-impact}, altering any attribute leads to a notable drop in effectiveness, confirming each one's critical role in strong black-box attacks.

\begin{figure}[t]
    \centering
    \includegraphics[width=\linewidth]{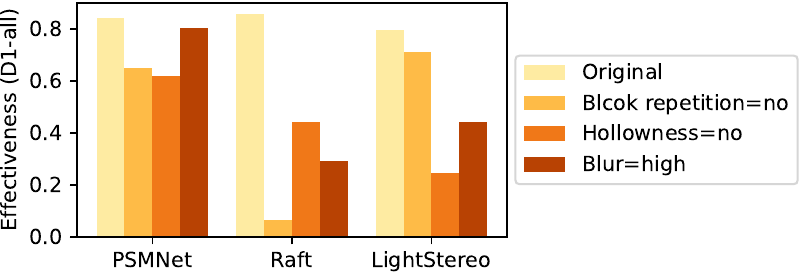}
    \caption{Individual impact of high-density hollowness, block repetition, and non-blur, on attack effectiveness.}
    \label{fig:individual-impact}
\end{figure}

\noindent\textbf{Analysis.} We attribute this effectiveness to the complementary disruptive effects of the three attributes.
\ding{172} \ul{\textit{Block repetition and high-density hollowness jointly impair the whole stereo matching pipeline}}. Block repetition primarily disrupts the early-stage matching by introducing multiple identical regions, severely increasing matching ambiguity. High-density hollowness violates smoothness priors in later stages, destabilizing cost aggregation and disparity estimation. This cross-stage disruption degrades both matching accuracy and disparity stability, highlighting the limited scope of current architectural advances (e.g., Mocha~\cite{chen2024mocha} and Monster~\cite{cheng2025monster}) in addressing vulnerabilities across all stages.
\ding{173} \ul{\textit{Sharp textures (non-blur) are essential to activating self-similarity effects.}} Blurring reduces edge clarity and repetitive cues, weakening the visual signals needed to induce ambiguity.

\begin{figure}[t]
    \centering
    \includegraphics[width=\linewidth]{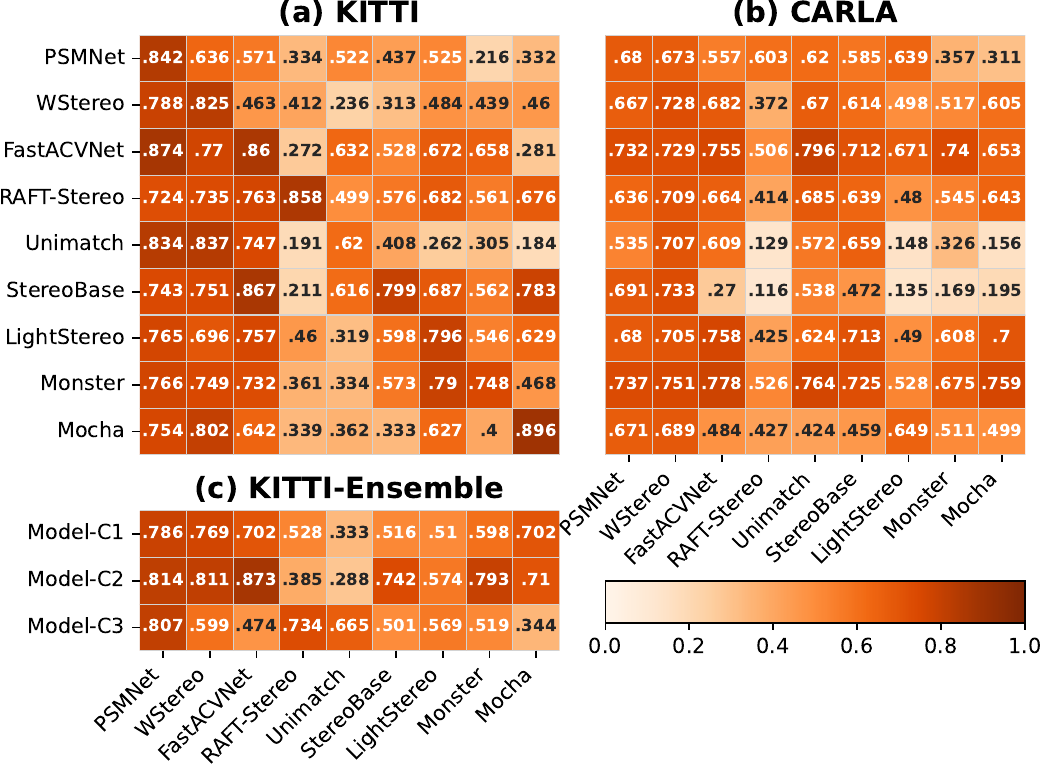}
    \caption{Evaluation results of \name on KITTI and CARLA. Each heatmap presents the effectiveness and transferability performance of discovered patterns, where rows correspond to different surrogate models or model combinations, and columns correspond to target models.  Model-C1: PSMNet, WStereo, and RAFT-Stereo. Model-C2: FastACVNet, Monster, and Light-Stereo. Model-C3: WStereo, Unimatch, and StereoBase.}
    \label{fig:trans-pattern}
\end{figure}

\noindent\textbf{Evaluations under \textit{non-interactive} attack.} We evaluate the transferability of all patterns in Table~\ref{tab:pattern-decomposition} by applying them to unseen SDE models. As shown in Figure~\ref{fig:trans-pattern}(a), all discovered patterns can constitute highly transferable adversarial patches. Table~\ref{tab:percent} further shows that \name consistently outperforms pixel-optimization-based methods.
Specifically, \name achieves lower $\mathcal{P}(<,\tau)$ at low thresholds ($\tau=0.1$, $0.2$, $0.3$), indicating stronger worst-case performance, and higher $\mathcal{P}(>,\tau)$ at high thresholds ($\tau=0.5$, $0.6$, $0.7$), reflecting improved best-case outcomes. These gains stem from structural similarities among the patterns, despite being independently optimized for different models.

\noindent\textbf{Model ensemble.} We apply a model ensemble strategy to \name and report results in Figure~\ref{fig:trans-pattern}(c). Patterns discovered with more surrogate models often show improved transferability. For instance, using only PSMNet yields the transferability of 0.332 to Mocha, while adding WStereo and RAFT-Stereo (Model-C1) boosts it to 0.702.

\begin{table}[t]
\centering
\caption{$\mathcal{P}(\circ,\tau)$ results for different attacks, aggregated from Figures~\ref{fig:usc} and~\ref{fig:trans-pattern}(a).}
\resizebox*{\linewidth}{!}{
\begin{tabular}{cc|ccc|ccc}
\hline
\multicolumn{2}{c|}{Method}      &  $<$0.1 ($\downarrow$)   &   $<$0.2 ($\downarrow$)  &   $<$0.3 ($\downarrow$)  &  $>$0.5 ($\uparrow$)   &  $>$0.6  ($\uparrow$)  &  $>$0.7  ($\uparrow$)  \\ \hline
\multirow{4}{*}{USC}   & Stage \ding{182} & 0.85 & 0.90 & 0.90 & 0.03 & 0.01 & 0.01 \\
                       & Stage \ding{183} & 0.92 & 0.94 & 0.99 & 0    & 0    & 0    \\
                       & Stage \ding{184} & 1    & 1    & 1    & 0    & 0    & 0    \\
                       & Stage \ding{185} & 1    & 1    & 1    & 0    & 0    & 0    \\ \hline
% \multirow{4}{*}{B-USC} & Stage \ding{182} & 0.19 & 0.36 & 0.44 & 0.40 & 0.29 & 0.19 \\
%                        & Stage \ding{183} & 0.22 & 0.39 & 0.46 & 0.29 & 0.18 & 0.07 \\
%                        & Stage \ding{184} & 0.28 & 0.38 & 0.42 & 0.26 & 0.18 & 0.13 \\
%                        & Stage \ding{185} & 0.29 & 0.43 & 0.50 & 0.33 & 0.22 & 0.10 \\ \hline
 \multicolumn{2}{c|}{\name} & \textbf{0} & \textbf{0.03}& \textbf{0.11}&\textbf{0.60}&\textbf{0.46}&\textbf{0.29} \\
  \hline
\end{tabular}
}
  \label{tab:percent}%
\end{table}

\subsection{Simulator Evaluation}
CARLA offers photorealistic scenes with fine-grained environmental control, allowing us to assess not only attack effectiveness and transferability, but also robustness under varying conditions like lighting and stereo baselines.

\noindent\textbf{Setup.} We control an ego vehicle in CARLA to collect data via its onboard stereo cameras. Following the KITTI setup, the stereo baseline is set to 0.54 meters (m), camera height to 1.65m, and FoV to $90^\circ$. We simulate four representative scenes: a straight urban street, a curved urban street, a suburban street, and a highway (Figure~\ref{fig:carlavis}). In each, a patch is placed 10m ahead, and the vehicle approaches until 2m. Closer distances are excluded as the patch would leave the camera's view. In each scene, we collect approximately 170 stereo image pairs at $1280\times 640$ resolution. We use a depth camera to capture corresponding depth maps and convert them to disparity maps via Eq.~(\ref{eq:stereo}), which serve as ground truth.

\noindent\textbf{Effectiveness.} Diagonal cells in Figure~\ref{fig:trans-pattern}(b) show high attack effectiveness in CARLA, with D1-all scores from 0.414 to 0.755 (averaged over all frames). Distance-specific results are provided in the robustness analysis.

\begin{figure}
    \centering
    \includegraphics[width=\linewidth]{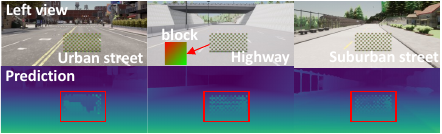}
    \caption{Transferable attack from WStereo to Mocha across different CARLA scenes at a distance of 7m. The predicted disparity maps exhibit pronounced deviations within the patch area, implying severe prediction error. }
    \label{fig:carlavis}
\end{figure}

\begin{figure}
    \centering
    \includegraphics[width=\linewidth]{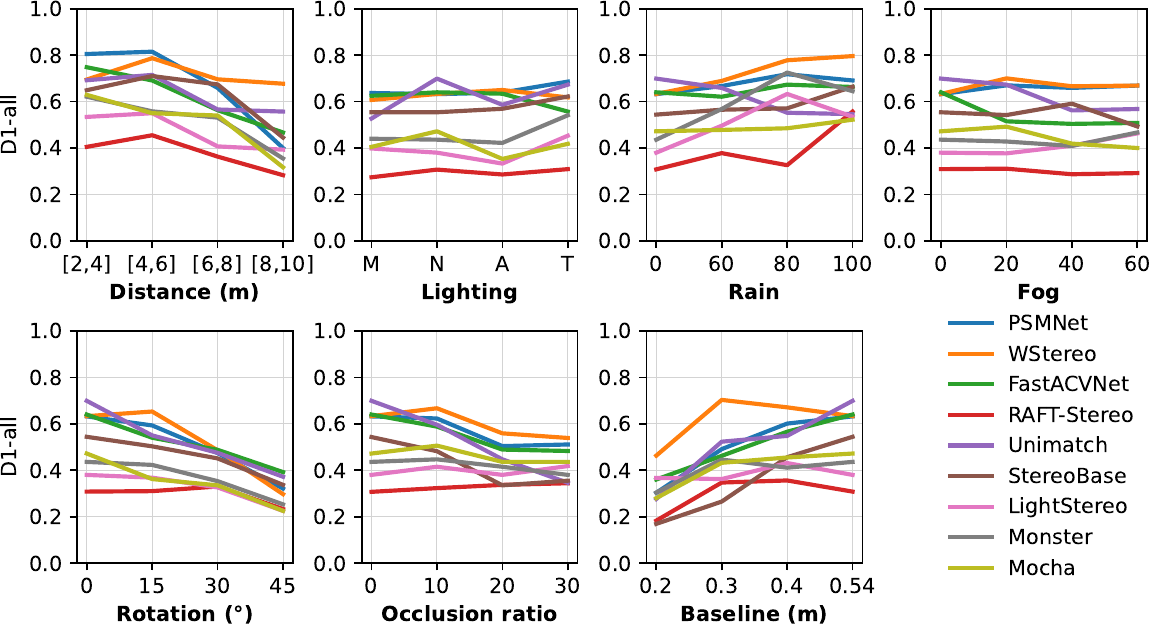}
    \caption{Robust evaluation on CARLA. In the distance plot, each x-axis value corresponds to a specific distance range. In the lighting plot, ``M'', ``N'', ``A'', and ``T'' mean morning, noon, afternoon, and twilight. For rain and fog, the x-axis indicates intensity levels. Occlusion is simulated by randomly masking a portion of the patch area, with x-axis values indicating the masking ratio.}
    \label{fig:calarrobust}
    % \vspace{-10pt}
\end{figure}

\noindent\textbf{Transferability.} Off-diagonal cells in Figure~\ref{fig:trans-pattern}(b) show \name's transferability in CARLA. The results are largely consistent with those on KITTI (Figure~\ref{fig:trans-pattern}(a)), as confirmed by Table~\ref{tab:percent}. This cross-dataset consistency highlights the strong generalization of our discovered patterns. Figure~\ref{fig:carlavis} further visualizes transferable attacks in different scenes.

\noindent\textbf{Robustness.} 
We assess \name's robustness in a CARLA suburban scene across 7 environmental factors, including distance, lighting, rain, fog, rotation, occlusion, and camera baseline, each with 4 variations (Figure~\ref{fig:calarrobust}). We observe that \name consistently performs well, with D1-all exceeding 0.2 even under long distances (8–10m), large rotations ($45^\circ$), and extreme weather. Notably, this robustness is achieved \textit{without any enhancement methods like EoT~\cite{athalye2018synthesizing}}, highlighting the inherent strength of our pattern-based design.

\subsection{Physical Evaluation} \label{sec:exp:phy}
Experiments on KITTI and CARLA show strong transferability and robustness of the discovered patterns, but digital results alone are insufficient. We conduct real-world tests to assess the sim-to-real effectiveness of \name fully.

\noindent\textbf{Setup.} 
We use a UGV equipped with industrial-grade cameras\footnote{\url{https://vitalvisiontechnology.com/alvium-g1-510c}} to collect real-world stereo images. The camera mounting positions are illustrated in Figure~\ref{fig:ugv}. The camera height (1.8m)  matches typical SUVs\footnote{\url{https://www.toyota.ca/en/vehicles/4runner/models-specifications}}, and the 0.3m baseline aligns with common stereo setups in real-world autonomous driving systems\footnote{\url{https://www.eetimes.com/ambarella-gives-robocars-double-vision}}. We use Mocha~\cite{chen2024mocha} and Monster~\cite{cheng2025monster} as victim models, as they demonstrate strong performance in physical-world scenarios. In each test, we place two patches 10m in front of the UGV for comparison: a pixel-optimization-driven patch $\mathbf{P}_{opt}$ and a pattern-driven patch $\mathbf{P}_{pat}$.
\begin{itemize}
\item $\mathbf{P}_{opt}$ is generated by attacking the feature extraction stage of WStereo using USC with EoT. Although this way achieves slightly better transferability than other settings in Figure~\ref{fig:usc}, its performance still lags far behind that of \name shown in Figure~\ref{fig:trans-pattern}.
% Considering that attacking WStereo's first stage achieves the best transferability among the USC settings in Figure~\ref{fig:usc} but still falls far behind \name, 
Therefore, we strengthen $\mathbf{P}_{opt}$ by introducing block repetition in this evaluation, making it a stronger baseline for comparison. Instead of optimizing the entire patch, we use USC to optimize a single block, and the rest of the patch is constructed by tiling this block. Figure~\ref{fig:busc} shows that this strategy substantially improves the attack performance of USC compared to Figure~\ref{fig:usc}.
% This setup extends the attack in~\cite{liu2024physical} by enhancing dynamic scene adaptation and selecting a more transferability-sensitive stage and a stronger surrogate model.
\item The basic appearance of $\mathbf{P}_{pat}$ is defined by the top three critical patterns (\textit{dense hollowness}, \textit{block repetition}, \textit{non-blur}) identified in earlier evaluations. We consider different texture blocks to individually compose $\mathbf{P}_{pat}$ (the bottom-right of Figure~\ref{fig:ugv}), enabling us to verify the effectiveness and robustness of these three key patterns.
\end{itemize}

\begin{figure}[t]
    \centering
    \includegraphics[width=\linewidth]{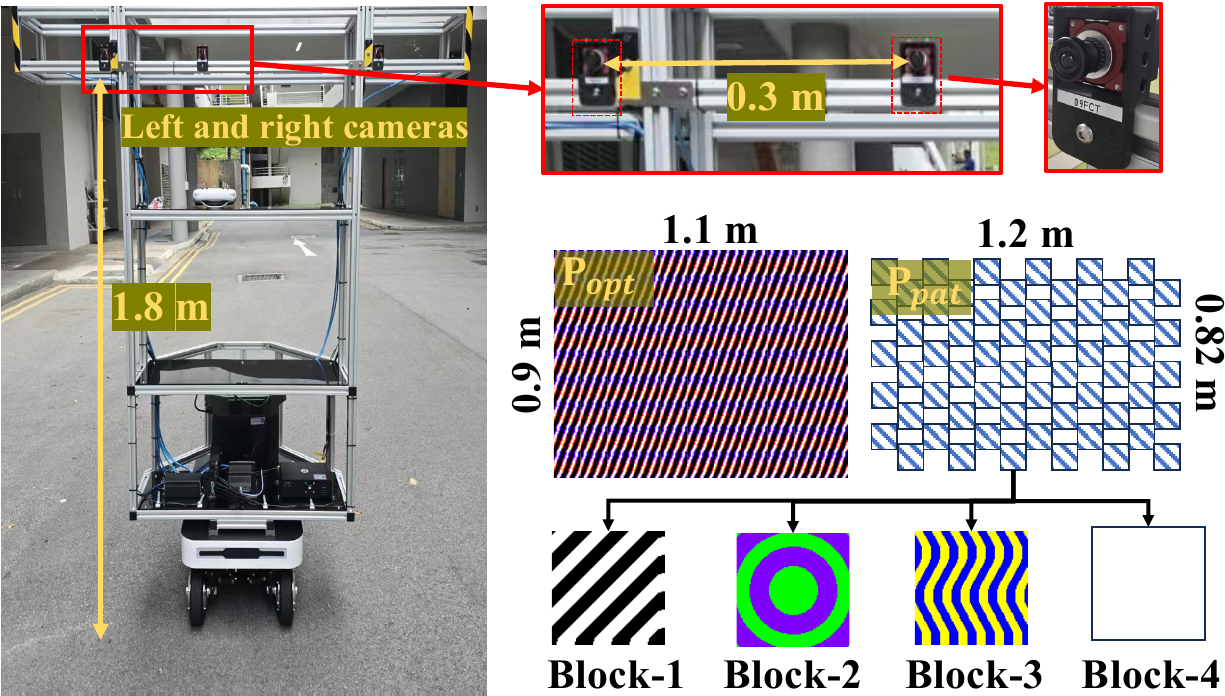}
    \caption{UGV platform and adversarial patches. $\mathbf{P}_{opt}$ is optimized using USC, while $\mathbf{P}_{pat}$ is pattern-driven, featuring dense hollowness, block repetition, and non-blur.}
    \label{fig:ugv}
    % \vspace{-10pt}
\end{figure}

\noindent Both $\mathbf{P}_{opt}$ and $\mathbf{P}_{pat}$ are composed of 0.1m$\times$0.1m blocks, and their full sizes are 0.9m$\times$1.1m and 0.82m$\times$1.2m, respectively, with roughly equal surface areas. The UGV approaches the two patches and stops at a distance of 2m, collecting approximately 144 stereo pairs per test. To mitigate positional bias, we repeat each test with the patches swapped. 

In the absence of dedicated depth sensors (e.g., LiDAR), we follow standard practice~\cite{liu2024physical} and adopt pseudo disparity as ground truth. As illustrated in  Figure~\ref{fig:phyvis}, we manually annotate regions of $\mathbf{P}_{opt}$ and $\mathbf{P}_{pat}$ on the collected images using Labelme\footnote{\url{https://labelme.io/}} and fill them with distinct colors (e.g., yellow and orange) that rarely appear in the environment. These masked images are then passed to the target model to generate the corresponding disparity maps (ground truth row of Figure~\ref{fig:phyvis}). The use of unique colors ensures reliable feature matching and accurate disparity estimation. Likewise, for $\mathbf{P}_{pat}$, we ignore its hollow structure when generating pseudo disparity to maintain estimation accuracy. As shown in the ground truth row, Mocha and Monster perform well within these masked regions. Finally, to exclude hollow areas of $\mathbf{P}_{pat}$ when computing D1-all, we also create a refined mask using Labelme (the bottom of Figure~\ref{fig:phyvis}).

% Finally, , we create refined masks for $\mathbf{P}_{pat}$ to exclude hollow areas (bottom row).

\subsubsection{Results \& Baseline Attack Comparisons} 

\begin{figure}
    \centering
    \includegraphics[width=\linewidth]{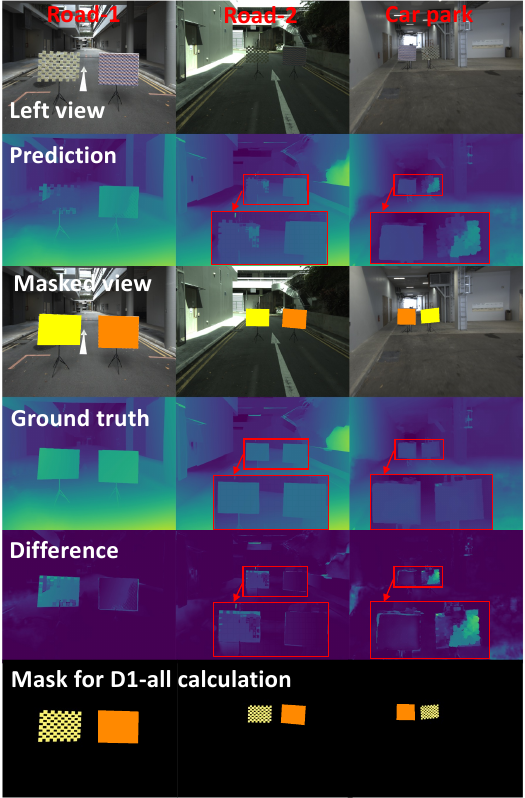}
    \caption{Comparison of pixel-optimization-based driven $\mathbf{P}_{opt}$ and pattern-driven patch $\mathbf{P}_{pat}$ in real-world tests. The first rows show the real-world inputs and predictions.
    In the third row, $\mathbf{P}_{pat}$ and $\mathbf{P}_{opt}$ regions are filled with yellow and orange, which are then used to produce pseudo disparity, serving as ground truth. Across various scenes, lighting conditions, and viewing distances, $\mathbf{P}_{pat}$ consistently causes greater disparity errors in the difference maps. The bottom row shows the refined masks used for the final D1-all computation, excluding the hollow regions of $\mathbf{P}_{pat}$. }
    \label{fig:phyvis}
    \vspace{-10pt}
\end{figure}

\begin{table*}[t]
  \centering
  \caption{Comparison of patch robustness with varying distances, scenes, lighting conditions, and rotations in the real world.}
    \resizebox*{\linewidth}{!}{
    \begin{tabular}{c|cccc|ccc|ccc|cccc}
    \hline
    \multirow{2}{*}{Patch} & \multicolumn{4}{c|}{Distance range}  & \multicolumn{3}{c|}{Lighting} & \multicolumn{3}{c|}{Scene}& \multicolumn{4}{c}{Rotation} \\
       & [2, 4] & [4, 6] & [6, 8] & [8, 10]  & \multicolumn{1}{c}{Low} & \multicolumn{1}{c}{Normal} & \multicolumn{1}{c|}{High} &  \multicolumn{1}{c}{Road-1} & \multicolumn{1}{c}{Road-2} & \multicolumn{1}{c|}{Car park}& \multicolumn{1}{c}{$-45^\circ$} & \multicolumn{1}{c}{$-30^\circ$} & \multicolumn{1}{c}{$+30^\circ$} & \multicolumn{1}{c}{$+45^\circ$}\\						
    \hline
    $\mathbf{P}_{opt}$  &     0.443   &     0.072   &   0.039    &    0.082     &  0.046 &  0.208  &  0.030  &    0.205   &   0.133   &  0.186   &     0.038   &  0.192    &   0.121 & 0.022 \\
    $\mathbf{P}_{pat}$  &   \textbf{0.907}    &   \textbf{0.672}   &   \textbf{0.373}    &    \textbf{0.223}   &   \textbf{0.428} &    \textbf{0.702}   &   \textbf{0.407}  & \textbf{0.719}  &  \textbf{0.770}  &  \textbf{0.639}   &  \textbf{0.547}     &     \textbf{0.606}  &    \textbf{0.643}   & \textbf{0.570}  \\ \hline
    \end{tabular}%
    }
  \label{tab:phycomparison}%
\end{table*}%
\begin{table*}[t]
  \centering
  \caption{Impact of the patch texture, patch size, and hollow scale on \name.}
  \resizebox{0.7\linewidth}{!}{
    \begin{tabular}{cccc|ccc|cc}
    \hline
    \multicolumn{4}{c|}{Texture block} & \multicolumn{3}{c|}{Patch size} & \multicolumn{2}{c}{Hollow scale}\\
    
    Block-1 & Block-2 & Block-3 & Block-4 & Large & Middle & Small & Dense & Sparse \\ \hline
    0.602  & 0.739 & 0.724 & 0.628 & 0.602  &  0.523   &  0.709  &  0.628    & 0.585 \\
    \hline
    \end{tabular}%
    }
  \label{tab:patchfactor}%
\end{table*}%

\noindent\textbf{Distance.}
% 3-1 & 3-2
% 4-1 & 4-2
We evaluate the performance of $\mathbf{P}_{pat}$ and $\mathbf{P}_{opt}$, under four distance intervals along a real-world road. Table~\ref{tab:phycomparison} reports the average results within each distance interval.
We observe that (i) $\mathbf{P}_{pat}$ consistently exhibits much better performance across the entire testing range than $\mathbf{P}_{opt}$, and (ii) $\mathbf{P}_{opt}$ is ineffective at long distances (e.g., $>$4m). 

\noindent\textbf{Lighting condition.} We evaluate the attack performance under three different lighting conditions in a road scene: (i) low-light (shadowed regions); (ii) normal-light (typical daylight); and (iii) high-light (direct noon sunlight).
As shown in Table~\ref{tab:phycomparison}, $\mathbf{P}_{pat}$ exhibits strong robustness to lighting variations and consistently outperforms $\mathbf{P}_{opt}$. 
Although \name's effectiveness decreases under extreme lighting conditions, such as low contrast in shadowed areas (middle column of Figure~\ref{fig:phyvis}), it still causes the victim model to produce erroneous predictions on over 40\% of the patch area.

\noindent\textbf{Scene.} We compare $\mathbf{P}_{pat}$ and $\mathbf{P}_{opt}$ in two road environments (Road-1 and Road-2) and one car park scenario (see Figure~\ref{fig:phyvis}).  
Table~\ref{tab:phycomparison} exhibits \name's strong generalization across diverse scenes. Although reduced parking lighting slightly affects the effectiveness of \name, it still achieves a D1 total score exceeding 0.67, demonstrating its robustness to lighting variations.

\noindent\textbf{Rotation.} We consider four rotation angles: $-45^\circ$, $-30^\circ$, $+30^\circ$, and $+45^\circ$, where the negative and positive signs indicate leftward and rightward rotations of the patch, respectively. As shown in Table~\ref{tab:phycomparison}, $\mathbf{P}_{pat}$ maintains its effectiveness across all tested rotation angles, while $\mathbf{P}_{opt}$ fails under large angle rotations, such as $\pm45^\circ$.

\noindent\textbf{Texture.} We apply different blocks shown in Figure~\ref{fig:ugv} to construct $\mathbf{P}_{opt}$ and evaluate their attack performance in road scenes. The results in Table~\ref{tab:patchfactor} show that patches characterized by dense hollowness, block repetition, and non-blur exhibit a certain degree of robustness to variations in texture.

\noindent\textbf{Patch size.} We consider two smaller versions of $\mathbf{P}_{pat}$ using block-1 as shown in Figure~\ref{fig:phyapp}. In Table~\ref{tab:patchfactor},  the large size is the default, while the medium (0.82m $\times$ 0.8m) and small (0.4m $\times$ 0.82m) sizes correspond to two-thirds and one-third of it.
Surprisingly, reducing the patch to one-third yields a higher D1-all value than the large size. We hypothesize two contributing factors:
(i) Small patches may blend into the background, offering fewer matching cues and leading to larger disparity errors, especially at long ranges where they occupy fewer pixels.
(ii) Since D1-all is computed only within the patch, smaller patches shrink the evaluation area, so even a few corrupted pixels can inflate the score.

\noindent\textbf{Hollow scale.} Table~\ref{tab:pattern-decomposition} favors a high-density hollow pattern for strong attack performance. To further validate its effect, we compare it with a sparse variant (the last column in Figure~\ref{fig:phyapp}). Table~\ref{tab:patchfactor} shows that both hollow patterns are effective, but the dense one performs better, likely because it induces sharper local disparity changes and more thoroughly disrupts the smoothness prior, which is a key assumption in SDE.

\section{Discussion} \label{sec:disscussion}

\noindent\textbf{Evaluations on conventional SDE methods.}
\name exploits shared weaknesses in the stereo matching pipeline, suggesting that its applicability is not limited to deep models but also extends to traditional non-learning-based methods. In Figure~\ref{fig:tradition}, we evaluate \name on three widely used conventional algorithms: SGBM~\cite{opencv_sgbm}, ELAS~\cite{geiger2010efficient}, and SGM~\cite{hirschmuller2007stereo}. The severe distortion in the predicted disparity within the patch regions highlights \name's generality across both learning-based and traditional SDE methods.

% Since \name targets inherent weaknesses in the stereo matching pipeline, it is also applicable to traditional non-learning-based methods.

\noindent\textbf{Potential defenses.} We begin with adversarial training (AT)\cite{cheng2023adversarial}, the most widely adopted defense strategy. We discuss two implementations: PGD-based AT and \name-based AT. In the inner loop of AT, the former iteratively optimizes patch pixels with PGD (e.g., 20 steps), while the latter trains an agent to search for pattern configurations. However, \name-based AT is not feasible for training, since agent optimization takes hundreds of steps (e.g., 800) per batch, which drastically slows down SDE training. So we focus on PGD-based AT and find that it fails to defeat \name (Appendix~\ref{sec:appendix:at}).

% Adversarial training~\cite{cheng2023adversarial} remains the most widely used defense, where the model is trained with adversarial examples generated at each step. However, directly integrating our patch generation method into the inner loop of adversarial training is impractical. Finding the optimal adversarial patterns using \name for each SDE model training step takes hundreds of steps, which would drastically slow down SDE training.

\begin{figure}[t]
    \centering
    \includegraphics[width=\linewidth]{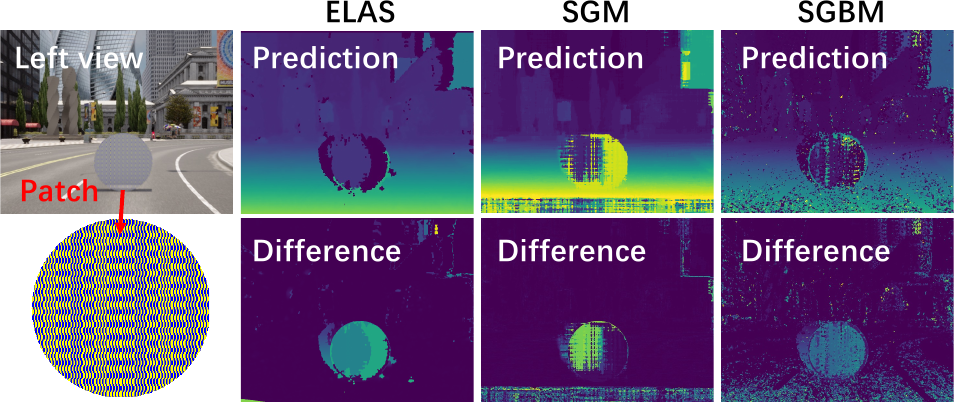}
    \caption{\name is also effective against traditional stereo matching methods. The patch is defined by the pattern configuration found for Mocha in Table~\ref{tab:pattern-decomposition}.}
    \label{fig:tradition}
\end{figure}

\begin{table}[t]
    \caption{Data augmentation-based defense. D1-all results of the fine-tuned models are reported for three input types: benign KITTI inputs (No), inputs patched with patterns discovered before fine-tuning (Old), and inputs patched with patterns re-discovered after fine-tuning (New).}
    \centering
    \begin{tabular}{c|ccc}
    \hline
    Pattern   & PSMNet & RAFT-Stereo & Mocha \\
           \hline
    No &  0.023 & 0.055 & 0.047 \\
    Old  & 0.019 & 0.009 & 0.052 \\
    New & 0.337 & 0.308 & 0.285 \\ \hline
    \end{tabular}

    \label{tab:defense}
\end{table}

Given the limitations of AT, we explore a data augmentation-based defense. At each training step, we generate patches by fixing the three most effective attributes (dense hollowness, block repetition, and non-blur) and randomly sampling the others. Using this strategy, we fine-tune PSMNet, RAFT-Stereo, and Mocha on the KITTI training set for 6000 steps with a learning rate of 0.001 and a batch size of 1. We then evaluate the fine-tuned models against both previously discovered patterns and new patterns generated by \name for the fine-tuned models. The results are summarized in Table~\ref{tab:defense}.~\ding{172}~\ul{\textit{Data augmentation provides strong robustness against known adversarial patterns.}} For fine-tuned models, the D1-all of previously discovered patterns drops below 0.1, compared to over 0.7 before fine-tuning.~\ding{173}~\ul{\textit{The defense shows limited generalization to unseen pattern configurations.}} Newly discovered patterns not covered by the augmentation set still induce non-negligible errors. For example, a chessboard-textured patch with sparse hollowness yields a D1-all of 0.337 on PSMNet. This evaluation indicates that even with full PPD coverage (Table~\ref{tab:primitive}) during data augmentation, adaptive attackers may still find vulnerabilities by extending the pattern space through additional primitives or richer options. 
% In Appendix~\ref{sec:appendix:at}, we also experiment with PGD-based adversarial training, which still fails to defend against \name. 
% Although the D1-all values of new patterns in Table~\ref{tab:defense} are generally lower than those in Table~\ref{tab:pattern-decomposition}, 
% Meanwhile, as the number of pattern configurations grows, incorporating them into data augmentation would make the fine-tuning increasingly difficult.}

% \lhc{We also experimented with PGD-based adversarial training in Appendix~\ref{sec:appendix:at}, but found that this way requires careful parameter tuning, and without it, training often collapses. Moreover, the resulting models still remain vulnerable to our pattern-based attacks even when training is stabilized.} 

Beyond single-modality defenses, one may also consider multi-sensor fusion, such as combining LiDAR with SDE. This way compensates for the sparsity of LiDAR while leveraging its direct 3D measurements as a reliable reference to correct errors in vision-based SDE. However, incorporating LiDAR inevitably increases the sensing workload and hardware costs, in contrast to the lightweight nature of SDE alone. In addition, since this work focuses on vulnerabilities in vision-based depth estimation, such a defense lies beyond our scope.

\noindent\textbf{Comparison with active sensor spoofing.} DoubleStar~\cite{zhou2022doublestar} proposes the only known active spoofing attack targeting SDE. It disrupts stereo matching by projecting asymmetric light into both cameras, creating false correspondences that violate epipolar constraints. Despite its effectiveness, DoubleStar suffers from several limitations. It requires real-time lighting projection, often involving human presence or external infrastructure. The attack is also susceptible to ambient lighting conditions and only works reliably in low-light environments. In contrast, \textit{\textbf{our method adopts an entirely different attack strategy}}. We design a passive patch-based attack that uses printable patches placed in the physical environment, which are naturally captured by the stereo cameras without requiring any active intervention. 

\noindent\textbf{Related attacks on MDE.} Previous works focus more on adversarial attacks against MDE models. Similar to the efforts on SDE~\cite{berger2022stereoscopic,liu2024physical}, these MDE attacks also rely on pixel-level optimization~\cite{wong2020targeted,cheng2022physical,guesmi2024saam,zheng2024physical} to distort depth predictions. In this work, we shift the focus to attacking SDE models, which are generally considered more robust and accurate due to their geometric consistency and reliance on binocular~\cite{smolyanskiy2018importance,tosi2019learning}. Hence, attacking SDE is much more challenging than MDE. 

\section{Conclusions}
In this paper, we conduct an in-depth investigation into the adversarial robustness of SDE under realistic constraints. We first present a unified framework to evaluate conventional pixel-optimization-based attacks across four stages of stereo matching, revealing their limited transferability despite access to gradients. Motivated by the discovery that structural vulnerabilities play a more critical role than fine-grained texture perturbations, we propose \name, the first pixel-optimization-free adversarial attack method targeting SDE models. It formulates patch generation as a search task over interpretable visual patterns. Extensive experiments across the KITTI dataset, CARLA simulator, and real-world scenarios demonstrate that \name achieves strong effectiveness and transferability in black-box settings, significantly outperforming existing pixel-optimization-based approaches.

\cleardoublepage
\bibliographystyle{plain}
\bibliography{sample}
\appendix

\section{Additional Details of USC and Our \name}

\subsection{USC}
We use a synthetic distribution $\mathcal{M}$ in Eq.(\ref{eq:loss2}) and Eq.(\ref{eq:loss3}) to guide patch generation when attacking the cost volume construction and aggregation stages. Specifically, $\mathcal{M}$ is defined using a sine wave, with two frequencies: $\frac{4}{192}$ and $\frac{16}{192}$, where 192 is the typical maximum disparity in SDE models. Figure~\ref{fig:synthetic} (first column) illustrates these two variants, with the top row showing the lower frequency and the bottom row the higher one. We generate patches based on both configurations, as shown in the last two columns of Figure~\ref{fig:synthetic}, but neither version demonstrates transferability in our experiments.

\subsection{\name} \label{sec:appendix:patchhunter}
\noindent\textbf{Pattern options.} 
For hollowness, we set hollow regions as $q \times q$ squares, with the spacing between adjacent squares also equal to $q$. $q=15$ for high density, while $q=30$ for low density. When block repetition is enabled, the block size is also set to $q \times q$. For brightness adjustment, we convert the image from the RGB color space to HSV, scale the V channel by a factor $\gamma$, and then convert the image back to RGB. We use $\gamma=1.2$, $1.0$, and $0.8$ to represent high, normal, and low brightness, respectively. For blurring, we apply the Gaussian blur function provided by OpenCV and vary the kernel size to control the severity. A kernel size of 5 corresponds to a slight blur, while a kernel size of 11 corresponds to a heavy blur. 

In the main text, we have discussed the impact of partial key pattern primitives, such as hollowness, blurriness, block repetition, and texture types. Table~\ref{tab:orientation} presents an ablation study focusing specifically on texture orientation. We adopt a base pattern composed of high-density hollowness, block repetition, and non-blur appearance. Then, we apply a consistent black-to-white gradient texture within each block to control orientation. The results demonstrate that changing only the orientation of the pattern while keeping all other factors fixed results in significant differences in attack effectiveness. This highlights the importance of treating orientation as an explicit and controllable parameter in the pattern space. Moreover, we observe that different SDE models exhibit varying sensitivity to specific orientations, suggesting model-specific vulnerabilities.

\noindent\textbf{Policy network.} We adopt a \textit{lightweight yet effective} network architecture based on a multi-head multilayer perceptron (MLP). The model consists of a shared bottom module and multiple independent heads. The shared module includes one embedding layer followed by two fully connected (FC) layers. The current option of each pattern primitive is first embedded into a 32-dimensional vector, resulting in a 9$\times$32 input (with 9 denoting the number of primitives). This is then transformed into a 9$\times$128 representation through the FC layers. Each head corresponds to a specific pattern primitive and contains two FC layers that map the flattened 1152-dimensional feature to a probability distribution over its candidate options. We apply greedy decoding by selecting the most probable option from each head to construct the complete action vector. Despite its simplicity, this architecture proves highly effective in practice. As shown in Figure~\ref{fig:ppo_curves}, the policy network converges to high reward within just 800 environment interaction steps, each corresponding to a complete observation–action–reward cycle. This is remarkably efficient given the large pattern space of nearly 44,000 unique configurations. Such efficiency stems from two key factors: (i) the highly structured pattern space defined by our PPD, which consists of semantically meaningful pattern options that have the potential to challenge various assumptions of SDE, and (ii) the simple yet effective design of our RL-based search procedure, which enables stable and fast convergence.

% the agent rapidly converges to strong adversarial pattern configurations with significantly improved rewards, validating the effectiveness of our network design.

\begin{figure}[t]
    \centering
    \includegraphics[width=\linewidth]{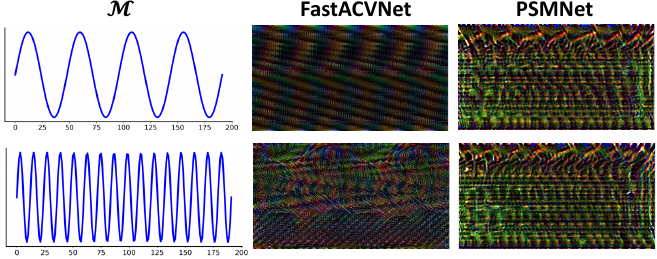}
    \caption{Multi-modal supervision $\mathcal{M}$ and corresponding Stage-\ding{183} patches on FastACVNet and PSMNet.}
    \label{fig:synthetic}
\end{figure}

\begin{table}[t]
    \centering
    \caption{Impact of orientation on gradient textures, measured by D1-all. The \textbf{bold values} mean the best attack effect, while the \ul{underline values} indicate the worst performance. The large discrepancy between the best and worst cases demonstrates that the orientation of gradient textures significantly impacts attack effectiveness.}
    \resizebox{\linewidth}{!}{
    \begin{tabular}{ccccc}
    \hline
        Orientation & FastACVNet & Unimatch & StereoBase & Mocha  \\ \hline
    $\rightarrow$&  \ul{0.233}& \textbf{0.486}& 0.221& 0.196\\
    $\leftarrow$& 0.436& 0.295& \ul{0.027}& \ul{0.185} \\
    $\uparrow$&  0.418& 0.203& 0.518& 0.366 \\
    $\downarrow$& 0.722& \ul{0.006}& \textbf{0.715}& 0.701 \\
    $\nearrow$& 0.605& 0.150& 0.100 & 0.366 \\
    $\searrow$& 0.672& 0.208&0.599&\textbf{0.788} \\
    $\swarrow$& \textbf{0.755}& 0.059& 0.144& 0.512 \\
    $\nwarrow$& 0.747& 0.116& 0.293& 0.302 \\ \hline
    \end{tabular}}
    \label{tab:orientation}
\end{table}

\begin{figure}[t]
    \centering
    \includegraphics[width=0.7\linewidth]{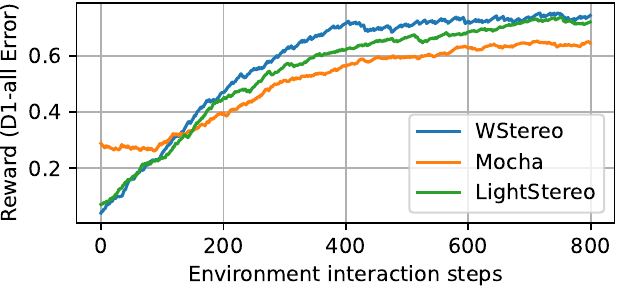}
    \caption{Reward convergence of the agent during training.}
    \label{fig:ppo_curves}
\end{figure}

% \begin{figure*}[t]
%     \centering
%     \includegraphics[width=\linewidth]{img/patches.pdf}
%     \caption{Visual examples of pattern primitives and their categorical options used in our patch design.}
%     \label{fig:opition}
% \end{figure*}

\begin{table*}[t]
    \centering
     \caption{Impact of patch size on USC. The source model is PSMNet, and the attack is applied to its first stage. The remaining columns represent unknown target models.}
    \resizebox{\linewidth}{!}{
    \begin{tabular}{c|ccccccccc}
    \hline
       Size & PSMNet& WStereo& FastACVNet&RAFT-Stereo&Unimatch&StereoBase&LightStereo&Monster&Mocha \\
        \hline
        100$\times$200&0.213&0.243&0.246&0.005&0.069&0.152&0.046&0.008&0.027\\ 
        150$\times$300&0.696&0.385&0.506&0.009&0.055&0.445&0.001&0.056&0.198 \\ \hline
    \end{tabular}
    }
   
    \label{tab:sizes}
\end{table*}

\noindent\textbf{PPO training.} In each epoch, we sample a trajectory by allowing the agent to interact with the environment for 16 time steps under the current policy. That trajectory consists of 16 state-action-reward tuples $(s_t, a_t, r_t)$, one for each time step $t$.
After that, we directly approximate the advantage function by normalizing the raw rewards collected within each epoch:
\begin{equation}
A_t = \frac{r_t - \bar{r}}{\sigma_r + \epsilon_1},
\end{equation}
where $A_t$ denotes the estimated advantage of taking action $a_t$ in state $s_t$ under the current policy, $\bar{r}$ and $\sigma_r$ are the mean and standard deviation of the rewards from the current epoch, and $\epsilon_1$ is a small constant (e.g., $1e^{-7}$). This normalization acts as a baseline and helps reduce the variance of policy gradient estimates. After that, we follow the standard PPO training procedure by optimizing the $\theta_P$ using the clipped surrogate objective:
\begin{equation}
\mathcal{L}(\theta_P) = \mathbb{E}_t \left[ \min\left( r_t(\theta_P) A_t,\ \text{clip}(r_t(\theta_P), 1 - \epsilon_2, 1 + \epsilon_2) A_t \right) \right],
\end{equation}
where $r_t(\theta_P) = \frac{\pi_{\theta_P}(a_t | s_t)}{\pi_{\theta_{P\text{old}}}(a_t | s_t)}$ is the probability ratio between the current and old policy. The clip function applies an element-wise constraint on $r_t(\theta_P)$, forcing each value to lie within the interval $[1 - \epsilon_2,\ 1 + \epsilon_2]$. $\epsilon_2$ (typically set to 0.1 or 0.2) controls the allowed deviation between the new and old policies, acting as a trust region to prevent overly aggressive updates. In experiments, we train the policy for up to 50 epochs using a learning rate of 0.001. As shown in Figure~\ref{fig:ppo_curves}, the agent’s reward increases rapidly and begins to stabilize after around 400 interaction steps. This demonstrates the high efficiency of the PPO-based search.

\begin{figure}[t]
    \centering
    \includegraphics[width=\linewidth]{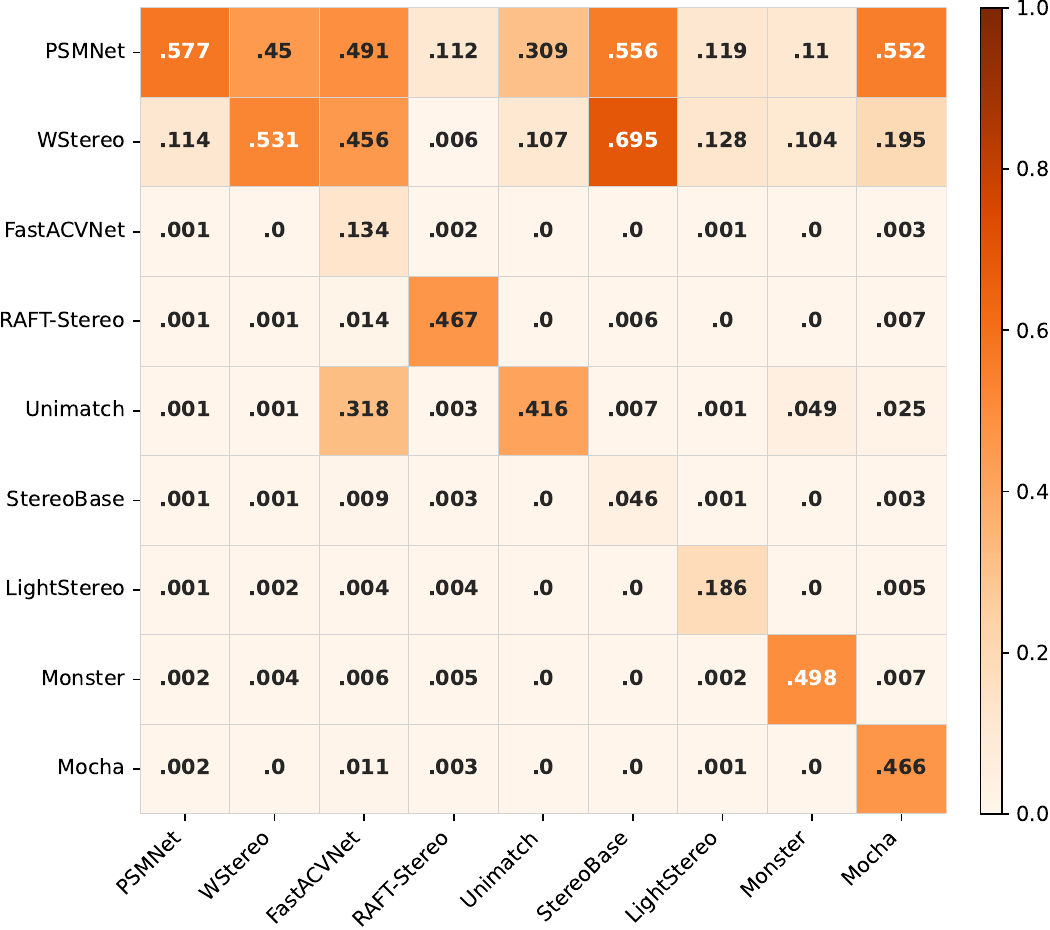}
    \caption{Attacking all four stages jointly using USC.}
    \label{fig:jointly}
\end{figure}

\begin{figure*}[t]
    \centering
    \includegraphics[width=\linewidth]{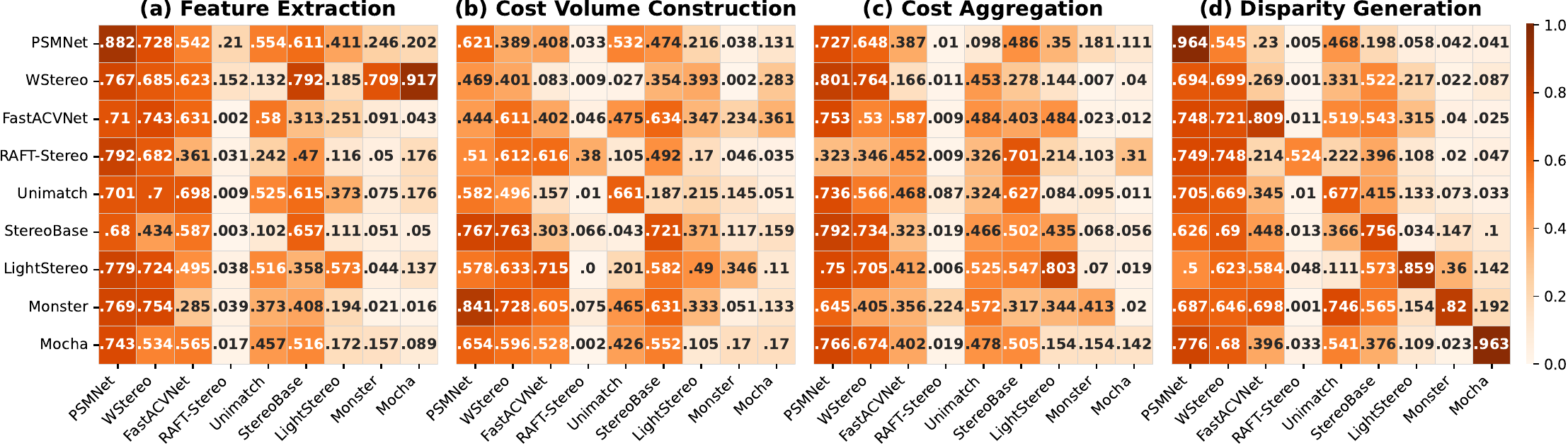}
    \caption{Enhanced USC with block repetition.}
    \label{fig:busc}
\end{figure*}

\begin{table*}[t]
\centering
\footnotesize
\caption{Adversarial patterns discovered by GA for PSMNet and LightStereo. ``---'' marks attributes not applicable to certain textures. The effectiveness means D1-all values.}
\resizebox*{\linewidth}{!}{
\begin{tabular}{c|ccccccccc|c}
\hline
{Target model} & {Patch shape} & {Hollowness} & {Block repetition} & {Brightness} & {Color} & {Texture type} & {Frequency} & {Orientation} & {Blur} & {Effectiveness} \\
\hline
PSMNet       & Circle   & High density & Yes & Normal  & Green             & Solid     & ---    & --- &  No    & 0.863  \\
% FastACVNet   & Circle& High density & Yes & High & White                & Solid     & --- & --- & No     & 0.860 \\
% StereoBase & Rectangle & No & No & Low & Green\&Purple&Gradient& Low& $\downarrow$ & No& 0.654 \\
LightStereo   & Rectangle& High density     & Yes & Normal  & Sky blue\&Light cyan            & Leopard      & Low    & ---        & No     & 0.667 \\
\hline
\end{tabular}
}
\label{tab:ga}
\end{table*}

\begin{table*}[t]
    \centering
     \caption{Transferability of patterns shown in Table~\ref{tab:ga}.}
    \resizebox{\linewidth}{!}{
    \begin{tabular}{c|ccccccccc}
    \hline
       Source model& PSMNet& WStereo& FastACVNet&RAFT-Stereo&Unimatch&StereoBase&LightStereo&Monster&Mocha \\
        \hline
        PSMNet&0.863 &0.735 &0.749 &0.340 &0.611 &0.327 &0.226 &0.114 &0.264\\ 
        LightStereo&0.750 &0.688 &0.708 &0.477 &0.402 &0.498 &0.667 &0.637 &0.584 \\ \hline
    \end{tabular}
    }
   
    \label{tab:ga_trans}
\end{table*}

\section{Additional Experiments} \label{sec:appendix:B}
\subsection{Impact of Patch Size on USC} \label{sec:appendix:patchsize}
We target the first stage of PSMNet and evaluate USC's transferability using different patch sizes. The evaluation results in Table~\ref{tab:sizes} indicate that reducing the patch size further degrades the already poor performance of USC.

\subsection{Multi-Stage Attacks via USC} \label{sec:appendix:multistage}
We have shown that USC fails to achieve high transferability when targeting any single stage of SDE (Figure~\ref{fig:usc}). We here extend the single-stage setting to a multi-stage attack, where the objective is reformulated as
\begin{equation}
    \mathcal{L}_{\theta_s} = \sum_{i=1}^4 \lambda^i\mathcal{L}^i_{\theta_s}.
\end{equation}
In experiments, we set each weight ($\lambda^i$) to 0.25 to treat all four stages equally, and keep all other hyperparameters consistent with Section~\ref{sec:investigation:setup}. Figure~\ref{fig:jointly} reports the D1-all results of this joint attack. Compared with Figure~\ref{fig:usc}(a), attacking all stages does not yield a clear improvement in transferability over attacking only the first stage.

\subsection{Improving USC via Block Repetition}
We investigate whether incorporating previously identified key patterns into USC can improve transferability. Specifically, we test block repetition: instead of directly optimizing the entire patch, USC optimizes a small block, which is then tiled to fill the whole patch. Figure~\ref{fig:busc} reports the attack performance in this setting, confirming that block repetition can substantially improve transferability compared to the original USC (Figure~\ref{fig:usc}). However, $\mathcal{P}(<,0.1)$ for the four stages in Figure~\ref{fig:busc} remain as high as 0.19, 0.22, 0.28, and 0.29, respectively, indicating that many source-target pairs still fail to transfer successfully. Therefore, the enhanced USC still struggles to achieve the strong transferability demonstrated by \name.

\subsection{Pattern Search with Genetic Algorithm} \label{sec:appendix:ga}
We further explore a representative heuristic search method, the Genetic Algorithm (GA), to search for patterns in the PPD-defined space. Specifically, we choose PSMNet and LightStereo as the source model and apply a standard GA algorithm to evolve candidate pattern combinations, including population initialization, crossover, and mutation. Table~\ref{tab:ga} reports the best-performing pattern combinations found by GA, which consistently involve the three key patterns (high-density hollowness, block repetition, and non-blur). Table~\ref{tab:ga_trans} shows the transferability of these patterns. This evaluation suggests that our attack framework is not restricted to a specific search strategy: both learning-based methods (e.g., PPO) and heuristic-based methods (e.g., GA) are applicable, while the core contribution lies in constructing the PPD and shifting the attack paradigm toward structured pattern-based search.

\begin{figure*}[t]
    \centering
    \includegraphics[width=\linewidth]{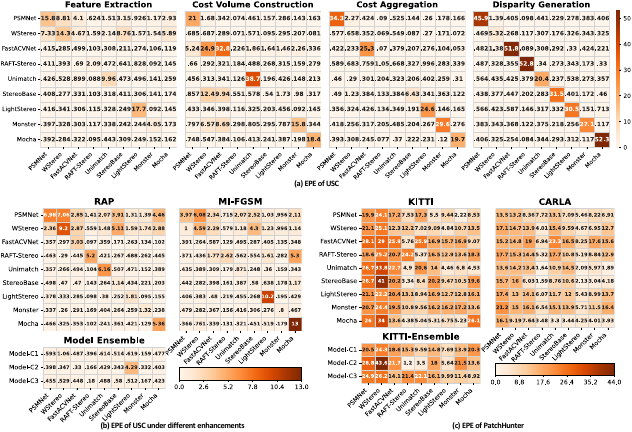}
    \caption{EPE Evaluation of Different Attacks. (a) and (b) correspond to evaluation results on KITTI of prior pixel-optimization-based attack methods, while (c) presents the evaluation of our pattern-based attack on both KITTI and CARLA.}
    \label{fig:epe}
    % \vspace{}
\end{figure*}
\subsection{Evaluations under Other Metrics} \label{sec:appendix:epe}
End-Point Error (EPE) is another popular metric in SDE studies~\cite{wong2021stereopagnosia,berger2022stereoscopic,wang2024left}, measuring the average absolute difference between the predicted disparity and the ground-truth disparity across all valid pixels in a stereo image. It is formally defined as 
\begin{equation}
\begin{split}
\text{EPE} &= \frac{1}{\left \| \Omega \right \| } \sum_{[h,w] \in \Omega }^{} \delta[h,w], \\
 \delta[h,w] &=  \left | \mathbf{\tilde{d}}[h,w] - \mathbf{d}[h,w] \right |. 
\end{split}
\end{equation}
Note that EPE is non-negative, whereas D1-all is constrained to values between 0 and 1.
We include EPE as a complementary metric to characterize the scale of disparity deviations, providing additional insights beyond the threshold-based D1-all metric. In Figure~\ref{fig:epe}(a) and (b), USC yields generally low EPE values in black-box settings, as indicated by many off-diagonal cells with values below 1. Such minimal perturbations to disparity predictions are unlikely to affect SDE performance in practice. In contrast, our pattern-driven \name achieves the best attack performance in black-box settings as it achieves EPE values above 3 across all possible transfer scenarios on the KITTI dataset (Figure~\ref{fig:epe}(c)). Notably, it induces high EPE results even on previously hard-to-attack models such as RAFT-Stereo, Mocha, and Monster. These observations are consistent with the D1-all-based evaluation in the main text, further confirming the superiority of \name in achieving both high attack effectiveness and strong transferability.

\subsection{PGD-based Adversarial Training} \label{sec:appendix:at} 
As discussed in Section~\ref{sec:disscussion}, directly incorporating \name into the adversarial training loop would be prohibitively slow. We here employ PGD~\cite{pgd}, a widely used and efficient pixel-level attack, for adversarial training and examine whether it can enhance robustness against pattern-based attacks. Specifically, at each training step, PGD generates an adversarial patch against the feature matching stage, with 20 iterations and a step size of 0.003. All other hyperparameters remain consistent with the augmentation-based setting (learning rate 0.001 and 6000 steps). However, PGD-based adversarial training often causes the model to fail. As shown in the first two columns of Figure~\ref{fig:at}, the predictions of PSMNet and RAFT-Stereo become unusable after adversarial training. We also tried some smaller learning rates (e.g., 0.0001) for these two models, but similar issues persisted. Notably, despite maintaining high performance on benign inputs after adversarial training (third column of Figure~\ref{fig:at}), Mocha remains vulnerable to the previously discovered pattern configurations (last column).

\subsection{Visualization examples of \name} \label{sec:appendix:visualization}
Figure~\ref{fig:carlaapp} and~\ref{fig:phyapp} present robustness tests of \name conducted in CARLA and the real world under varying environmental conditions. In all tested conditions, the patch region shows clear highlights in the difference maps, indicating significant disparity prediction errors caused by the adversarial patch.

\begin{figure*}[t]
    \centering
    \includegraphics[width=\linewidth]{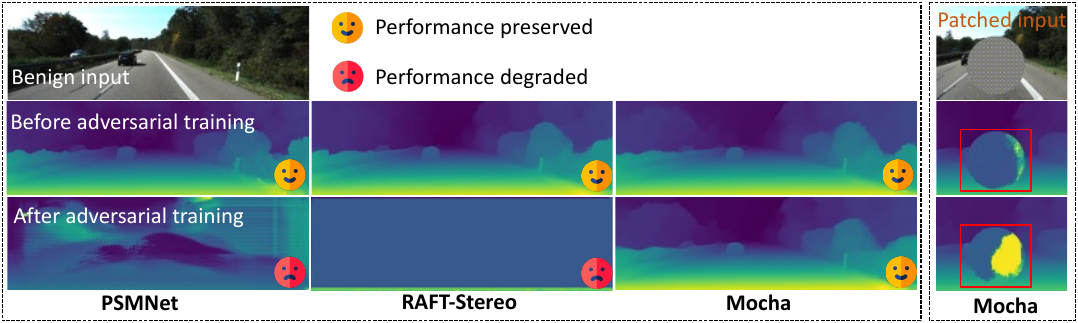}
    \caption{PGD-based adversarial training on different SDE models.}
    \label{fig:at}
\end{figure*}

\begin{figure*}
    \centering
    \includegraphics[width=\linewidth]{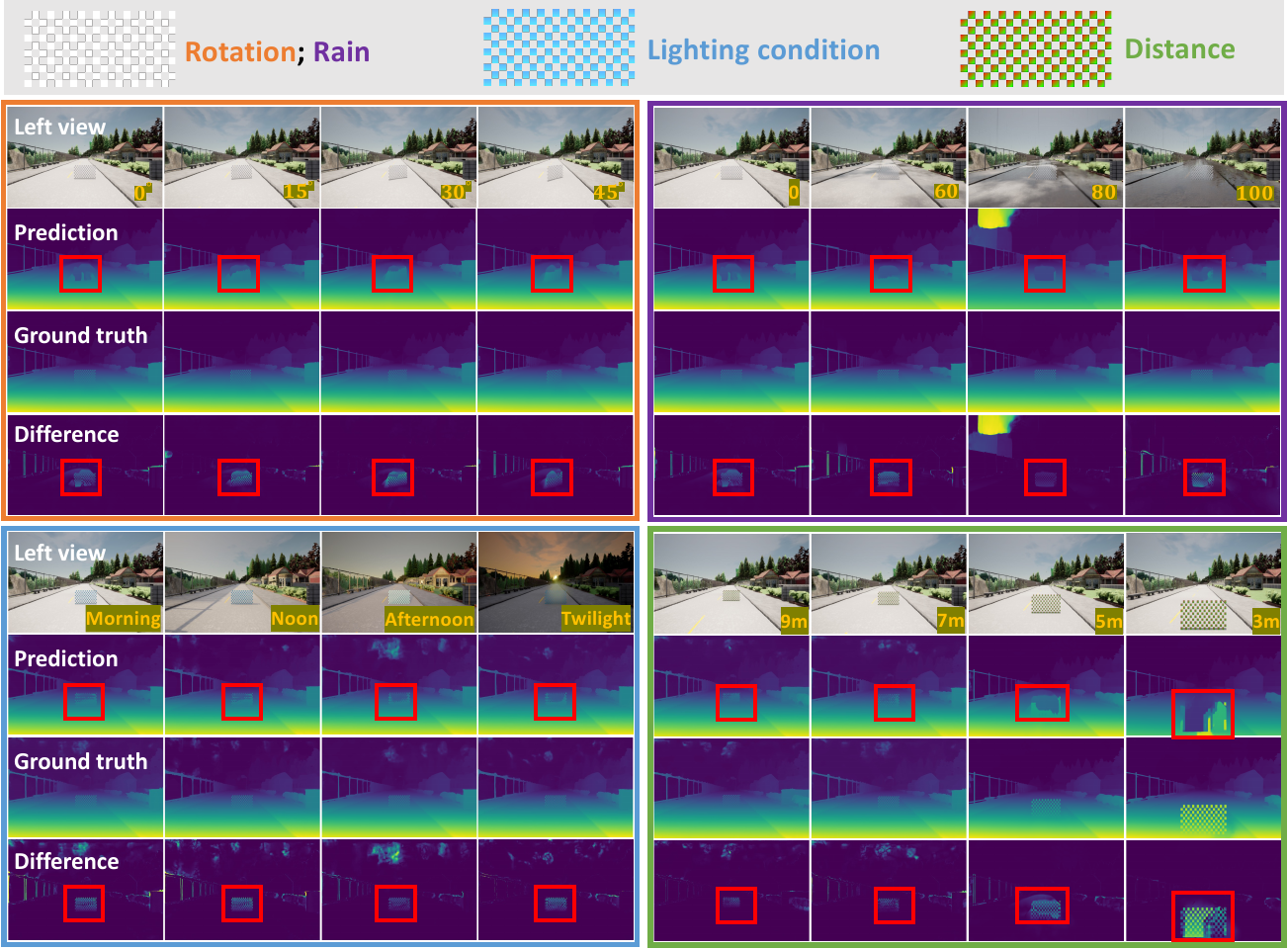}
    \caption{Robustness evaluation of \name under varying environmental conditions in CARLA. Each set of subfigures is outlined in a color that corresponds to the category label above: orange for rotation, purple for rain, blue for lighting condition, and green for distance.}
    \label{fig:carlaapp}
\end{figure*}
\begin{figure*}
     \centering
    \includegraphics[width=\linewidth]{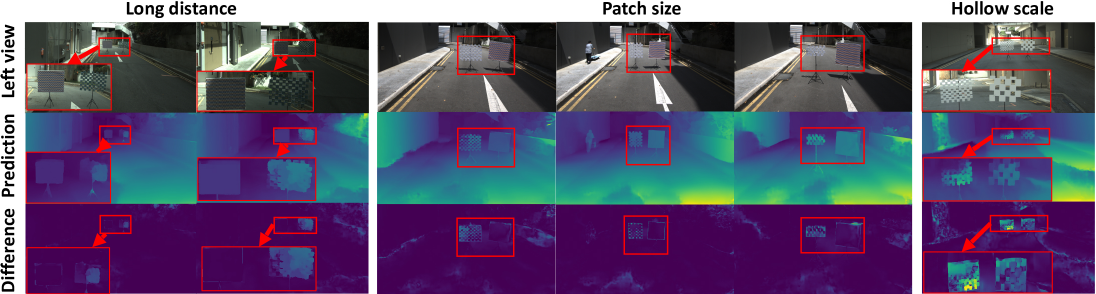}
    \caption{Robustness evaluation of \name under varying environmental conditions in the real world.}
    \label{fig:phyapp}
\end{figure*}

%%%%%%%%%%%%%%%%%%%%%%%%%%%%%%%%%%%%%%%%%%%%%%%%%%%%%%%%%%%%%%%%%%%%%%%%%%%%%%%%
\end{document}